\documentclass[11pt]{article}

\usepackage[final]{acl}

\usepackage{times}
\usepackage{latexsym}

\usepackage[T1]{fontenc}

\usepackage[utf8]{inputenc}

\usepackage{microtype}

\usepackage{inconsolata}

\usepackage{graphicx}
\usepackage{tabularx}
\usepackage{enumitem}
\usepackage{array}
\usepackage{ragged2e}
\usepackage{enumerate}
\usepackage[justification=raggedright]{caption}
\usepackage{floatflt}
\usepackage{booktabs}
\usepackage{multirow}
\usepackage[normalem]{ulem}
\useunder{\uline}{\ul}{}
\usepackage{makecell}
\usepackage{listings}
\usepackage{tcolorbox}
\tcbuselibrary{skins,breakable}
\usepackage{xcolor}
\usepackage{natbib}
\usepackage{amsmath}      
\usepackage{amssymb}      
\usepackage{bm}           

\usepackage[ruled,vlined,linesnumbered]{algorithm2e}

\newtcolorbox{custombox}[1]{
  colback=lightgray!5, 
  colframe=gray!85!black, 
  arc=3pt, 
  title={\bfseries #1}, 
  fonttitle=\large, 
  boxrule=1.5pt, 
  top=2pt, 
  bottom=2pt, 
  left=10pt, 
  right=10pt, 
  breakable, 
  enhanced
}

\title{Beyond Compliance: A Resistance-Informed Motivation Reasoning Framework for Challenging Psychological Client Simulation}

\author{
    Danni Liu\textsuperscript{1}, 
    Bo Liu\textsuperscript{1,2}\thanks{Corresponding author. Email: bliu@seu.edu.cn}, 
    Yuxin Hu\textsuperscript{1}, 
    Hantao Zhao\textsuperscript{1}, 
    Yan Liu\textsuperscript{3},
    Ding Ding\textsuperscript{1},
    Jiahui Jin\textsuperscript{1},
    Jiuxin Cao\textsuperscript{1,2} \\
    \textsuperscript{1} Southeast University, Nanjing, China \\
    \textsuperscript{2} Purple Mountain Laboratories, Nanjing, China \\
    \textsuperscript{3} The Nanjing Derong Wisdom Information Technology Co., Ltd., Nanjing, China \\
    \{danniliu, bliu, yuxinhu, jx.cao\}@seu.edu.cn, yanliu@drzh-atmr.cn
}

\begin{document}
\maketitle
\begin{abstract}
Psychological client simulators have emerged as a scalable solution for training and evaluating counselor trainees and psychological LLMs. Yet existing simulators exhibit unrealistic over-compliance, leaving counselors underprepared for the challenging behaviors common in real-world practice. 
To bridge this gap, we present ResistClient, which systematically models challenging client behaviors grounded in Client Resistance Theory by integrating external behaviors with underlying motivational mechanisms. To this end, we propose Resistance-Informed Motivation Reasoning (RIMR), a two-stage training framework. First, RIMR mitigates compliance bias via supervised fine-tuning on RPC, a large-scale resistance-oriented psychological conversation dataset covering diverse client profiles. Second, beyond surface-level response imitation, RIMR models psychologically coherent motivation reasoning before response generation, jointly optimizing motivation authenticity and response consistency via process-supervised reinforcement learning. Extensive automatic and expert evaluations show that ResistClient substantially outperforms existing simulators in challenge fidelity, behavioral plausibility, and reasoning coherence. Moreover, ResistClient facilities evaluation of psychological LLMs under challenging conditions, offering new optimization directions for mental health dialogue systems.


\end{abstract}

\section{Introduction}
Mental health disorders affect over one billion people worldwide, yet access to qualified counselors remains severely limited \cite{world2025}. Motivated by this limitation, large language models (LLMs) have increasingly been explored for counselor training \cite{steenstra2025scaffolding}, evaluation \cite{xiao2025mentrasuite}, and supervision \cite{lin2023supervisorbot}, where client simulators play a central role by providing scalable, repeatable practice environments \cite{wang2024patient}. However, existing psychological client simulators often exhibit over-compliant behaviors compared to real clients \cite{yang2025consistent}, manifesting as unusually high openness \cite{kim2025can}, excessive receptivity, and positive emotional stability\cite{ming2025annaagent}. This behavioral misalignment leads to the under-exploration of challenging therapeutic scenarios, ultimately constraining the effectiveness and reliability of both counselor trainees and psychological LLMs in realistic clinical contexts.

\begin{figure}
    \centering
    \includegraphics[width=0.95\linewidth]{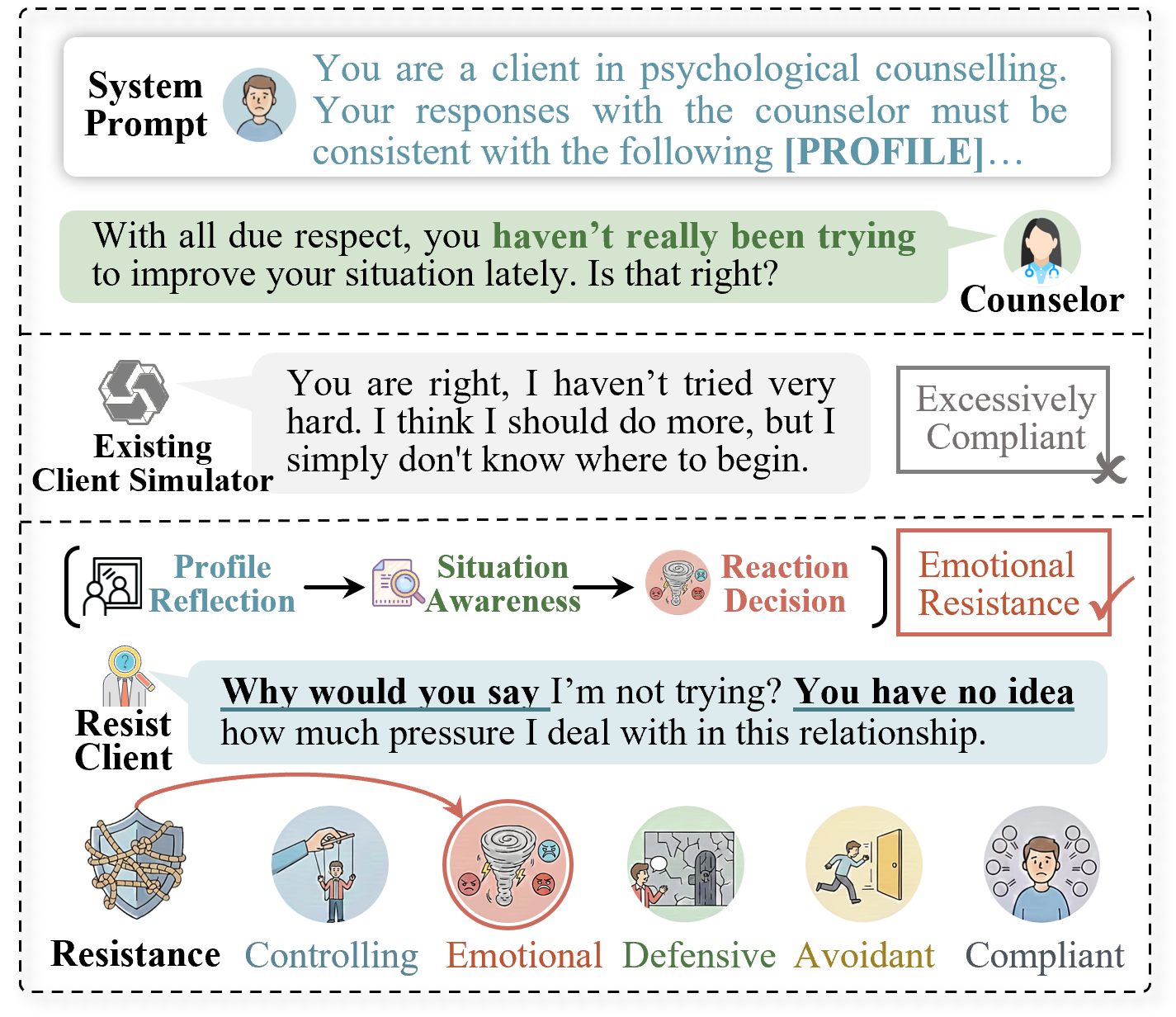}
    \caption{Existing LLM-based client simulators exhibit overly-compliance. ResistClient generates authentic resistant behaviors through motivation reasoning steps.}
    \label{fig:example}
\end{figure}

In real therapeutic practice, counselors frequently encounter "difficult" or "challenging" clients who exhibit resistance—defined as behaviors that avoid or reduce self-disclosure requested by the counselor because such communication causes discomfort or anxiety \cite{otani1989client}. Resistance is both inevitable and consequential: virtually all clients exhibit some degree of resistance during counseling, and extensive clinical literature shows that the effectiveness of resistance management critically shapes therapeutic processes and outcomes \cite{strean1985resolving}. Consequently, explicitly incorporating resistance mechanisms into client simulation is essential for narrowing the mismatch between idealized, compliant simulations and the challenging realities of clinical practice.

This paper aims to develop a resistance-informed client simulator that authentically replicates the challenging dynamics of real-world therapeutic interactions. Two key challenges arise:

\noindent 1) \textbf{Unrealistic challenging behaviors induced by compliance bias}: Existing client simulators rely on profile-conditioned generation using pre-aligned LLMs \cite{wang2024patient}, whose inherent compliant bias fundamentally conflicts with clinical non-compliance \cite{wang2025know}. Lacking exposure to authentic resistance during pre-training, prompt-level adjustments \cite{yang2025consistent,ming2025annaagent,kim2025can} yield only superficial difficulty, failing to replicate the challenging interactions and psychological depth characteristic of real-world therapeutic encounters.

\noindent 2) \textbf{Shallow simulation induced by absence of internal mechanisms}: Psychological theory conceptualizes resistance as cognitive–affective processes \cite{otani1989client, chamberlain1984observation}, not isolated behaviors. In contrast, existing client simulators focus on direct response generation without modeling the motivational mechanisms underlying these behaviors. Consequently, simulated reactions are reduced to surface-level response fitting, obscuring whether the behaviors reflect psychologically coherent processes or merely replicate superficial response patterns.

To address these challenges, we propose ResistClient (Figure~\ref{fig:example}), a challenging client simulator that replicates therapeutic resistance through explicit motivation reasoning. ResistClient is instantiated via Resistance-Informed Motivation Reasoning (RIMR), a two-stage training framework:
(1) Resistance-Informed Supervised Fine-Tuning mitigates the compliant bias in pre-trained LLMs. We construct the Resistance-Informed Psychological Conversations (RPC) dataset, which conceptualizes diverse client profiles from real conversations and augments challenging behaviors grounded in Client Resistance Theory \cite{chamberlain1984observation}. Fine-tuning on RPC enables the model to learn expression patterns of different resistance types from real interactions, recalibrating its behavioral distribution toward realistic non-compliance.
(2) Motivation Reasoning Reinforcement Learning (MRRL) addresses the lack of internal mechanism modeling by encouraging explicit motivation reasoning before response generation. Mirroring psychological processes of real clients, the model generates responses through structured reasoning steps, integrating profile-based cognitive reflection and situational awareness. Using process-supervised rewards derived from expert feedback, MRRL optimizes the step-wise authenticity of motivation reasoning and its consistency with the generated responses, enabling psychologically coherent behaviors.
Together, RIMR enables ResistClient to balance challenge intensity with behavioral plausibility, supporting realistic and interpretable simulation of difficult therapeutic interactions.

Our contributions can be summarized as follows: 

\begin{itemize}
\setlength{\itemsep}{-3pt}
\item To our best knowledge, this is the first work to systematically investigate and model challenging behaviors in client simulation. We propose ResistClient, shifting the prevailing paradigm from over-compliance to authentic challenging interactions.

\item We introduce a novel training mechanism RIMR and RPC dataset. Moving beyond shallow responses imitation, our method generates psychologically coherent behaviors through motivation reasoning, enabling high-fidelity simulation across diverse client profiles.

\item Extensive automatic and expert evaluations demonstrate ResistClient achieves state-of-the-art challenge intensity and behavioral plausibility. By replicating realistic challenging interactions, our work provides a critical new perspective for evaluating clinical reliability of psychological LLMs.
\end{itemize}

\section{Related Work}
Client simulators struggle with over-compliance, while existing building paradigms prioritize external responses over internal psychological processes. We tackle this by adapting reasoning reinforcement learning (RRL) grounded in psychological resistance theory, aligning both behavioral patterns and internal cognition with real challenging clients. 

\noindent\textbf{Overly-Compliant Behavior in Client Simulation}.
Recent LLM-based client simulators enable profile-conditioned generation \cite{wang2024patient, chen2025socialsim}, extensions improving consistency via state tracking or memory mechanisms \cite{yang2025consistent, ming2025annaagent}. Despite prompt-level mitigation (e.g., information withholding \cite{kim2025can}, low receptivity setting \cite{yang2025consistent}, or emotion injection \cite{ming2025annaagent}), existing simulators remain overly compliant due to the inherent compliant bias of pre-aligned LLMs \cite{lu2024large, xu2023baize}. More critically, psychological theory views challenging client behaviors as manifestations of underlying cognitive–affective resistance \cite{otani1989client}, rather than isolated behavior. ResistClient addresses these by grounding simulation in Client Resistance Theory \cite{chamberlain1984observation}, jointly modeling resistant behaviors and their underlying motivation reasoning.

\noindent\textbf{Building Paradigms for User Simulation}.
Existing user simulators fall into four paradigms: (1) \textit{Profile-conditioned generation} \cite{wang2024patient} relies on intrinsic LLM reasoning but struggles with behavioral consistency. (2) \textit{Agent-based user simulators} \cite{ming2025annaagent, yang2025consistent} employ agentic frameworks to improve consistency but lack utterance-level realism due to limited real-data training \cite{wang2025know}. (3) \textit{Conditional supervised fine-tuning} \cite{kong2024platolm, wang2025know} learns from real data but only achieves surface-level imitation. (4)  \textit{Non-collaborative user simulation} counteracts the inherent politeness of LLMs by modeling uncooperative behaviors \cite{zhang2024strength, xiao2025stimulate}, but lacks specific expertise in the psychological field. Across all paradigms, existing methods prioritize observable behavior over latent reasoning, limiting their utility for psychological training.

\noindent\textbf{Reasoning Reinforcement Learning for LLMs.}
Recent RRL frameworks (e.g., DeepSeek-R1 \cite{guo2025deepseek}) explicitly model intermediate reasoning steps, using RL to align reasoning quality with human preferences. Recent efforts extend this paradigm to psychological domains: Psyche-R1 \cite{dai2025psyche} enhances empathy and expertise reasoning, while Mindora \cite{xiao2025mentrasuite} imporves clinically aligned counselor reasoning. However, they target supporter-side cognition, leaving client-side motivation reasoning unexplored. We pioneer RRL for client simulation, adapting GRPO with process-level rewards to ensure step-wise validity and reasoning–response consistency.

\section{ResistClient Construction}
\begin{figure*}
    \centering
     \includegraphics[width=0.95\linewidth]{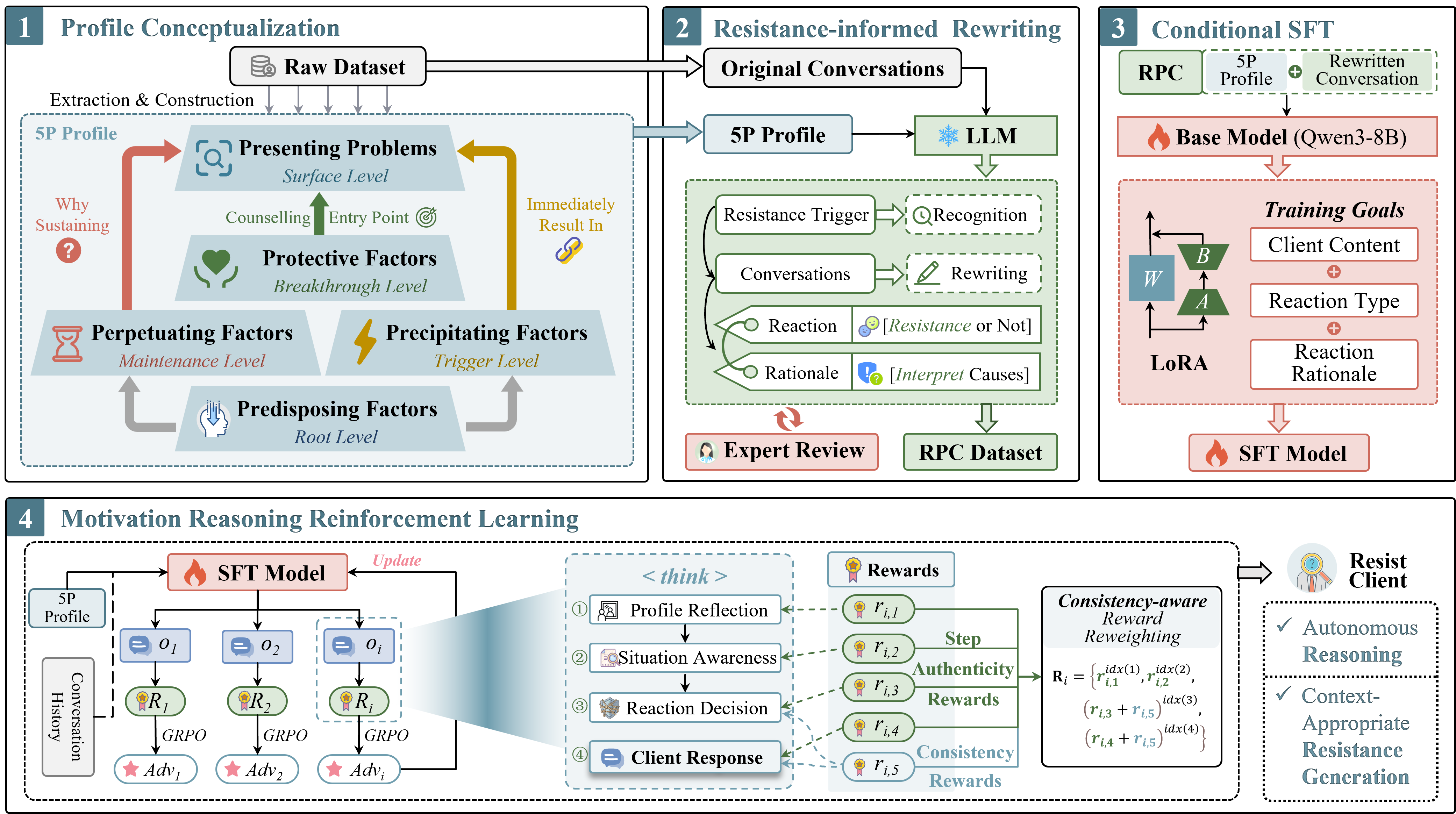}
    \caption{Overview of our proposed ResistClient with Resistance-informed Motivation Reasoning framework.}
    \label{fig:framework}
\end{figure*}

To address the over-compliance bias in existing client simulators, we propose RIMR, a two-stage training framework for psychologically grounded resistance behaviors. As illustrated in Figure \ref{fig:framework}, RIMR consists of: (1) Resistance-Informed Supervised Fine-Tuning on our RPC dataset, and (2) Motivation Reasoning Reinforcement Learning.

\subsection{RPC Dataset Construction}
Existing psychological conversation datasets exhibit compliant bias, where clients respond unrealistically compliantly even to potentially resistance-eliciting counselor utterances.
To mitigate this bias, we construct the RPC dataset through (i) psychologically grounded profile conceptualization and (ii) resistance-informed conversation rewriting.
\subsubsection{Profile Conceptualization from Psychological Conversation}
Rather than relying on shallow personas, we adopt a clinically grounded Client Profile Schema inspired by the 5P case formulation model \cite{johnstone2013formulation}, which links internal psychological states to observable reactions through: (1) Presenting Problems: primary concerns; (2) Predisposing Factors: historical vulnerabilities; (3) Precipitating Factors: recent triggers; (4) Perpetuating Factors: maintaining circumstances; and (5) Protective Factors: recovery resources. This causally-informed schema ensures client responses manifests as a psychologically grounded process rather than random occurrences.

To ensure the authenticity and diversity of simulated interactions, we build profiles utilizing ProPsyC, a large-scale repository of real-world psychological conversations \cite{hu2025psyadvisor}. Following the schema above, we employ DeepSeek-V3.2 to extract structured 5P profiles with few-shot guidance from expert-annotated examples. Profile quality is evaluated along coverage and faithfulness dimensions following \citet{wang2025know}, with low-consistency samples filtered out (Appendix~\ref{sec:a}).

\subsubsection{Conversation Rewriting with Resistance Reactions} 
Although resistance is pervasive in counseling practice \cite{strean1985resolving}, it is rarely reflected in open-source datasets due to ethical constraints and privacy-preserving curation that systematically suppress challenging reactions. To mitigate this distributional bias while preserving the overall coherence and intent of the original conversations, we propose a Resistance-Informed Conversation Rewriting framework to introduce psychologically grounded resistance reactions at contextually appropriate moments, through controlled, theory-driven rewriting. Our framework is described as follows:

\noindent\textbf{Resistance Trigger Recognition}. Drawing on interviews with counselors, we identify resistance triggers as counselor interventions that are likely to elicit client resistance when interacting with clients’ psychological vulnerabilities (Table \ref{tab:trigger}). Rather than relying on surface cues alone, trigger recognition is conditioned on both the local conversational context and high-risk features in the client’s 5P profile. During rewriting, each counselor turn is examined for such trigger characteristics in a context-sensitive and profile-aware manner. 

\noindent\textbf{Resistance-Informed Rewriting}. Guided by Client Resistance Theory \citep{chamberlain1984observation}, we define a comprehensive reaction taxonomy consisting of five resistance types (Controlling, Emotional, Defensive, Avoidant, and Compliant) and two cooperative types (Non-resistant and Facilitative) (Table~\ref{tab:label-types}). When a resistance trigger is detected, the subsequent client response is rewritten to reflect the most contextually appropriate reaction type. To prevent semantic drift, rewriting is strictly localized: only the immediate response and up to three subsequent turns are modified, with at most one primary resistance episode per session.

\noindent\textbf{Client Reaction Annotation}. Each client response is annotated with a reaction label and a brief reaction rationale explaining its underlying motivation. To balance scalability and validity, we adopt LLM-assisted annotation with expert verification. Few-shot prompts with curated examples guide initial labeling, while professional counselors verify and refine the annotations to ensure clinical validity.

Using DeepSeek-V3.2 and this framework, we construct the RPC from ProPsyC. As shown in Appendix \ref{sec:a}, the resulting dataset contains 1,849 complete counseling sessions, spanning 14 common counseling topics, with 1,761 sessions exhibiting resistance behaviors. Each session is paired with a validated 5P client profile and thoroughly annotated conversation, providing a high-quality foundation for training resistance-aware client simulators.

\subsection{Conditional Supervised Fine-Tuning} To mitigate the over-compliance inherited from pretraining distributions, our model performs profile-conditioned supervised fine-tuning on RPC. This stage enables the model to learn structured mappings from client profiles and interaction contexts to specific behavioral outputs—including reaction types, responses, and underlying reasons—as conceptualized in Client Resistance Theory. Formally, given a client profile $p$, conversation history $\mathcal{H}_{t-1}$, and counselor utterance $u^c_t$, the model $\pi_{\theta_{\text{sft}}}$ is trained to predict a client response tuple $u^a_t$ (reaction type, response, reason). The supervised fine-tuning objective minimizes the following loss:
\begin{equation}
\mathcal{L}_{sft} = -\sum_{t=1}^{T} \log P(u^a_{t,j} | u^a_{t,<j},u^c_{t},\mathcal{H}_{t-1},p)
\end{equation}
where $T$ is the total number of conversation turns and $u^a_{t,j}$ denotes the $j$-th token of the $u^a_t$.

\subsection{Motivation Reasoning RL}
Generating realistic client reactions is an inherently open-ended task, where surface-level language fitting is insufficient to ensure psychological fidelity. While supervised fine-tuning addresses surface-level over-compliance, realistic resistance behaviors further require explicit modeling of the underlying motivation reasoning process. To this end, we propose MRRL, which aligns structured client-side psychological reasoning with human preferences via process-supervised reinforcement learning.

\subsubsection{Structured Motivation Reasoning Generation and Annotation}
\noindent\textbf{Motivation Reasoning Structure.} We decompose the motivation reasoning process into three structured steps: 1) \textit{Profile Reflection}, which reflects potential resistance tendencies by integrating stable cognitive and emotional factors from the 5P profile; 2) \textit{Situation Awareness}, which analyzes conversation history and current counselor utterance to infer the client’s momentary psychological state; and 3) \textit{Reaction Decision}, specifying the reaction type and expected behavioral characteristics. This structured process mirrors human-like psychological reasoning, enabling both interpretability and improved realism.

\noindent\textbf{Process-Supervised Reward Annotation.} We evaluate the motivation reasoning process and client response generation from two perspectives. First, we assess the step-wise quality of each reasoning step and the final response to ensure psychological plausibility. Second, to optimize the model's ability to maintain coherence between the reaction type decided through reasoning and the final utterance, we evaluate reasoning-response consistency. Crucially, to align the reasoning process with human judgment, we employ expert-annotated process-level rewards on sampled outputs from the SFT model $\pi_{\theta_{\text{sft}}}$. Detailed annotation protocols are provided in the supplementary materials \ref{sec:c}.

\subsubsection{RL with Consistency-aware Reward Reweighting}
To jointly optimize reasoning validity and reasoning–response alignment, we introduce a consistency-aware reward reweighting strategy within GRPO. Based on the expert-annotated samples, we employ GRPO in an offline setting with process-supervised rewards. Specifically, for each context $q$ (including client profile and conversation history), GRPO samples a group of outputs $\{o_i\}_{i=1}^G$ from the old policy $\pi_{\theta sft}$ and optimizes the policy model by maximizing:
\begin{equation} 
\begin{aligned}
    \mathcal{J}_{GR}&_{PO}(\theta) = \mathbb{E}\left[{q \sim P(Q), \{o_i\}_{i=1}^G \sim \pi_{\theta_{\text{sft}}}(O|q)}\right] \\
    & \frac{1}{G} \sum_{i=1}^G \frac{1}{|o_i|} \sum_{t=1}^{|o_i|} \bigg\{ \text{min} \bigg[ \frac{\pi_\theta(o_{i,t}|q,o_{i,<t})}{\pi_{\theta_{\text{sft}}}(o_{i,t}|q,o_{i,<t})}\hat{A}_{i,t},  \\
    & \text{clip}\left( \frac{\pi_\theta(o_{i,t}|q,o_{i,<t})}{\pi_{\theta_{\text{sft}}}(o_{i,t}|q,o_{i,<t})},1-\epsilon,1+\epsilon \right)\hat{A}_{i,t} \bigg] \\
    & - \beta \mathbb{D}_{KL}(\pi_\theta || \pi_{ref}) \bigg\}
\end{aligned}
\end{equation}
where $o_{i,t}$ represents the $t$-th token in output $o_i$, $\epsilon$ is the clipping parameter, and $\beta$ controls the KL divergence penalty. To enforce the alignment between reasoning and response, we utilize a Consistency-aware Reward Reweighting strategy. For each sampled output $o_i$, we have step-wise rewards $r_{i,1}$ (Reflection), $r_{i,2}$ (Awareness), $r_{i,3}$ (Decision), and $r_{i,4}$ (Response). Additionally, a consistency reward $r_{i,5}$ is applied to both the decision and response steps to enforce semantic alignment. As shown in Figure \ref{fig:framework}, the reward vector $\mathbf{R}_i$ for each step index $k$ is constructed as:
\begin{equation}
\begin{aligned}
\mathbf{R}_i = &\{r_{i,1}^{\mathrm{idx}(1)}, r_{i,2}^{\mathrm{idx}(2)}, (r_{i,3} + r_{i,5})^{\mathrm{idx}(3)}, \\&(r_{i,4} + r_{i,5})^{\mathrm{idx}(4)}\}
\end{aligned}
\end{equation}
where $\mathrm{idx}(k)$ refers to the end token of reasoning step $k$. These step-level rewards are normalized across the sampled group:
$\tilde{r}_{i}^{\mathrm{idx}(k)} = \frac{r_i^{\mathrm{idx}(k)} - \text{mean}(\mathbf{R})}{\text{std}(\mathbf{R})}$. Subsequently, the advantage of each token is calculated as the sum of normalized rewards from all following steps:
$\hat{A}_{i,t} = \sum_{\mathrm{idx}(k) \geq t} \widetilde{r}_{i}^{\mathrm{idx}(k)}$, and the policy is optimized by maximizing the objective defined above. As a result, the consistency signal influences both decision and response steps and is back-propagated to earlier reasoning tokens, explicitly coupling internal reasoning with its surface realization and encouraging coherent and psychologically consistent client behaviors. Full algorithmic details are provided in Algorithm~\ref{alg:grpo}.

\begin{figure}
    \centering
    \includegraphics[width=0.8\linewidth]{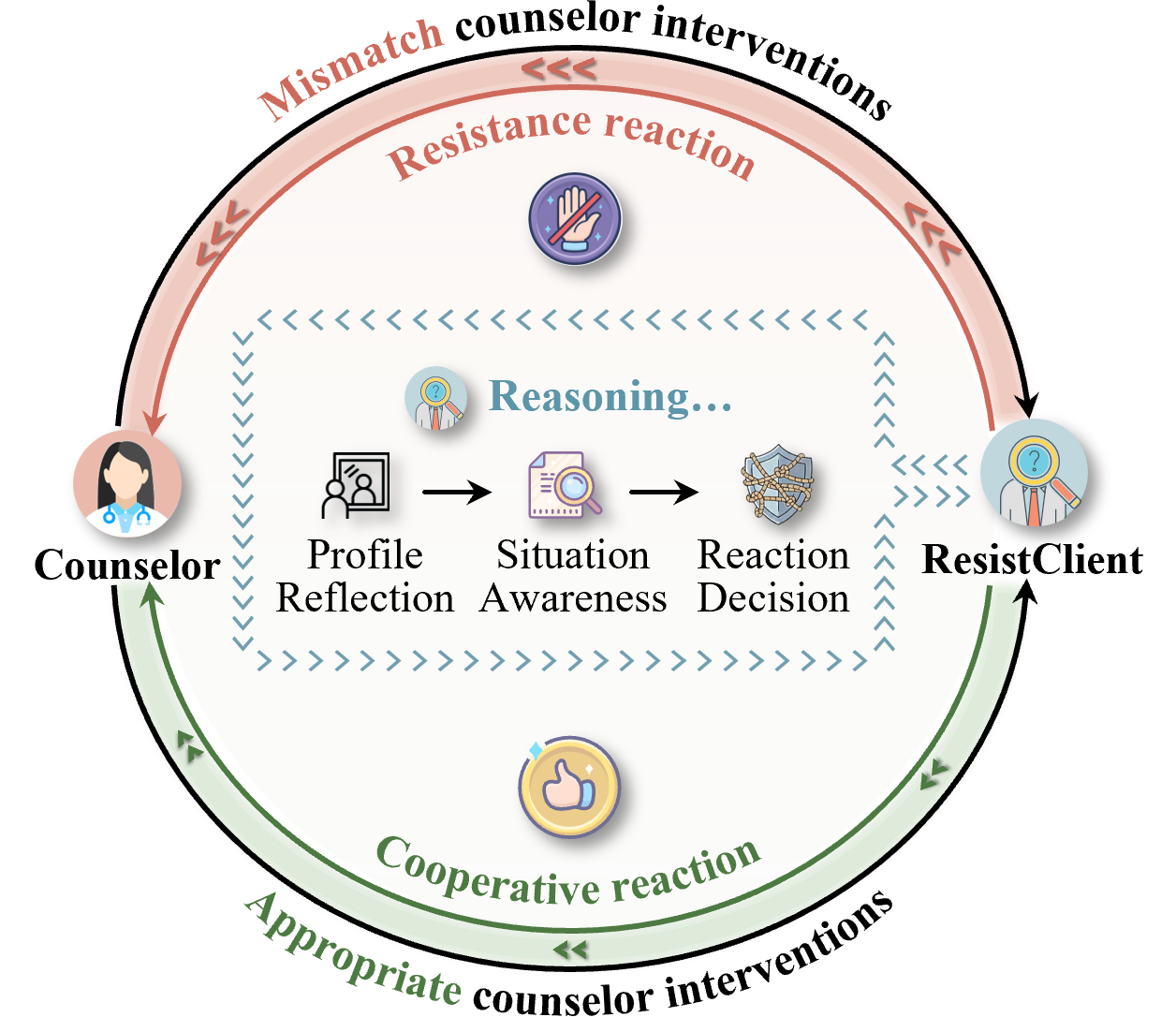}
    \caption{Application of ResistClient in interactive counseling training. }
    \label{fig:client-use}
\end{figure}

Through this two-stage approach, ResistClient effectively captures complex client reaction patterns with high psychological reasoning fidelity. As shown in Figure~\ref{fig:client-use}, the simulator provides human-like feedback—both cooperative and resistant—conditioned on the appropriateness of counseling strategies, enabling more realistic and informative interactive training for counselors and psychological LLMs.

\section{Experiments}
To rigorously evaluate ResistClient, we conduct comprehensive experiments to assess its effectiveness, realism, and practical utility in simulating challenging clients. In particular, we aim to answer the following research questions (RQs): 
\textbf{RQ1}: How effectively can ResistClient simulate client resistance behaviors compared with reasoning LLM baselines?
\textbf{RQ2}: How do the two training stages contribute to overall simulation performance?
\textbf{RQ3}: How realistic are the challenging behaviors generated by ResistClient compared with existing approaches?
\textbf{RQ4}: What is the performance of current psychological LLMs when handling resistance in interactions with ResistClient?

\subsection{Resistance Simulation Capability Analysis}
\subsubsection{Experiment Setup}
\noindent\textbf{Evaluation framework}. This experiment evaluates models’ ability to simulate client resistance under controlled counseling scenarios by reusing counselor turns from the original dataset. Client simulators are initialized with 100 randomly sampled profiles from RPC. All baselines receive identical prompt descriptions of resistance and cooperative reactions to ensure fairness (Appendix~ \ref{sec:d-prompt}). For reliability, we generate 3 sessions per profile.

\begin{table*}[!t]
\centering
\setlength{\tabcolsep}{7pt}
\renewcommand{\arraystretch}{1}
\begin{tabular}{lcccccccc}
\toprule[1.2pt]
\multirow{2}{*}{\centering \textbf{Models}\vspace{-7pt}} & \multicolumn{4}{c}{\textbf{Automated Metrics(\%)}} & \multicolumn{3}{c}{\textbf{Manual Metrics}} \\
\cmidrule(lr){2-5} \cmidrule(lr){6-8}
& Precision & Recall & F1 Score & RTF & Fid. ($\uparrow$) & Rat. ($\uparrow$) & Qua. ($\uparrow$)\\
\midrule[0.8pt]
GPT-5.1 & 59.31 & 62.88 & 61.04 & 35.94 & 1.42 & 1.35 & 2.52 \\
DeepSeek-V3.2 & 52.87 & 57.56 & 55.12 & 36.91 & 1.29 & 1.21 & 2.40 \\
Kimi-K2-thinking & 47.08 & 51.19 & 49.05 & 36.86 & 1.24 & 1.19 & 2.18 \\
GLM-4.6 & 49.10 & 54.72 & 51.76 & 37.78 & 1.27 & 1.23 & 2.34 \\
\cline{1-8}
Qwen3-8B & 36.52 & 48.54 & 41.68 & 45.06 & 1.10 & 1.04 & 1.88 \\
DeepSeek-R1-8B & 34.67 & 46.26 & 39.64 & \uline{\textbf{45.23}} & 0.98 & 1.08 & 1.72 \\
\cline{1-8}
Qwen3-8B-SFT & 63.54 & 73.90 & 68.33 & 39.43 & 1.46 & 1.41 & 2.39 \\
ResistClient & \uline{\textbf{70.38}} & \uline{\textbf{78.95}} & \uline{\textbf{74.42}} & 38.03 & \uline{\textbf{1.63}} & \uline{\textbf{1.58}} & \uline{\textbf{2.61}} \\
\bottomrule[1.2pt]
\end{tabular}
\caption{Resistance simulation performance of different models in psychological conversations.}
\label{tab:resistance}
\end{table*}

\noindent\textbf{Baselines}. We compare ResistClient with representative reasoning-capable LLMs, including: 1) large-scale models: GPT-5.1 \cite{openai2025gpt51}, DeepSeek-V3.2\cite{liu2025deepseekv3}, Kimi-K2-thinking\cite{team2025kimi}, GLM-4.6\cite{zeng2025glm}, 2) open-source small-scale models: Qwen3-8B\cite{yang2025qwen3}, DeepSeek-R1-8B\cite{guo2025deepseek}, and 3) our ablations: Qwen3-8B-SFT.

\noindent\textbf{Evaluation Metrics}. Automated metrics evaluate the timing and accuracy of resistance generation, including Precision, Recall, F1, and Resistance Trigger Frequency (RTF). Human evaluation (0–3 scale) assesses psychological quality across: Resistance Fidelity (Fid.), measuring alignment between reasoning reaction type and generated response; Resistance Rationality (Rat.), evaluating contextual appropriateness of resistance; and Reasoning Quality (Qua.), assessing the coherence and plausibility of underlying motivation reasoning. All human scores are averaged over expert annotators (Appendix~ \ref{sec:d-metrics1}).

\noindent\textbf{Implementation Details}. We use Qwen3-8B as the backbone model for its strong reasoning capability and instruction-following performance. The model is first fine-tuned on RPC using CSFT for 2 epochs on an NVIDIA A100 (80GB). Subsequently, we apply the MRRL stage for another 2 epochs to align the reasoning process with human clinical preferences. Inference uses $T=0.7$, $top\_p=0.8$, and $top\_k=20$.

\subsubsection{Performance Comparison (RQ1)}

Table~\ref{tab:resistance} shows that ResistClient consistently outperforms all baselines across automated evaluations, indicating improved resistance timing and simulation accuracy. Specifically, the superior precision indicates ResistClient avoids excessive or inappropriate resistance, while the high recall shows it effectively recognizes conflict situations without over-compliance. Smaller open-source models (e.g., Qwen3-8B, DeepSeek-R1-8B) exhibit higher RTF but substantially lower Precision, indicating a tendency toward over-triggering resistance. Human evaluation confirms that ResistClient achieves the highest scores in fidelity, rationality, and reasoning quality, indicating that both its surface behaviors and internal motivation reasoning are psychologically grounded and contextually appropriate. A detailed case study illustrating these dynamics is provided in the Appendix~\ref{sec:e}.

\subsubsection{Ablation Study (RQ2)}
To examine the contribution of each training stage, we compare ResistClient with two variants: 1) Qwen3-8B, using prompt-based conditioning only, and 2) Qwen3-8B-SFT, trained with CSFT on RPC. Figure \ref{fig:client-use} visualizes reaction-type alignment using confusion matrices, where rows correspond to ground-truth reactions and columns to generated reactions. The diagonal mass increases from the prompt-only to SFT and to the full framework, indicating progressively improved simulation accuracy across reaction categories and confirming the contributions of both training stages. Notably, prompt-only model shows a strong bias toward cooperative reactions, consistent with the over-compliance tendency of pretrained LLMs. SFT effectively reduces this bias by learning diverse resistance patterns from RPC. MRRL further improves performance by explicitly modeling motivation reasoning and aligning with human preferences, particularly reducing confusion among resistance types. Furthermore, the competitive performance of the SFT-only variant against large closed-source models highlights the effectiveness of the RPC dataset.

\begin{figure}
    \includegraphics[width=0.95\linewidth]{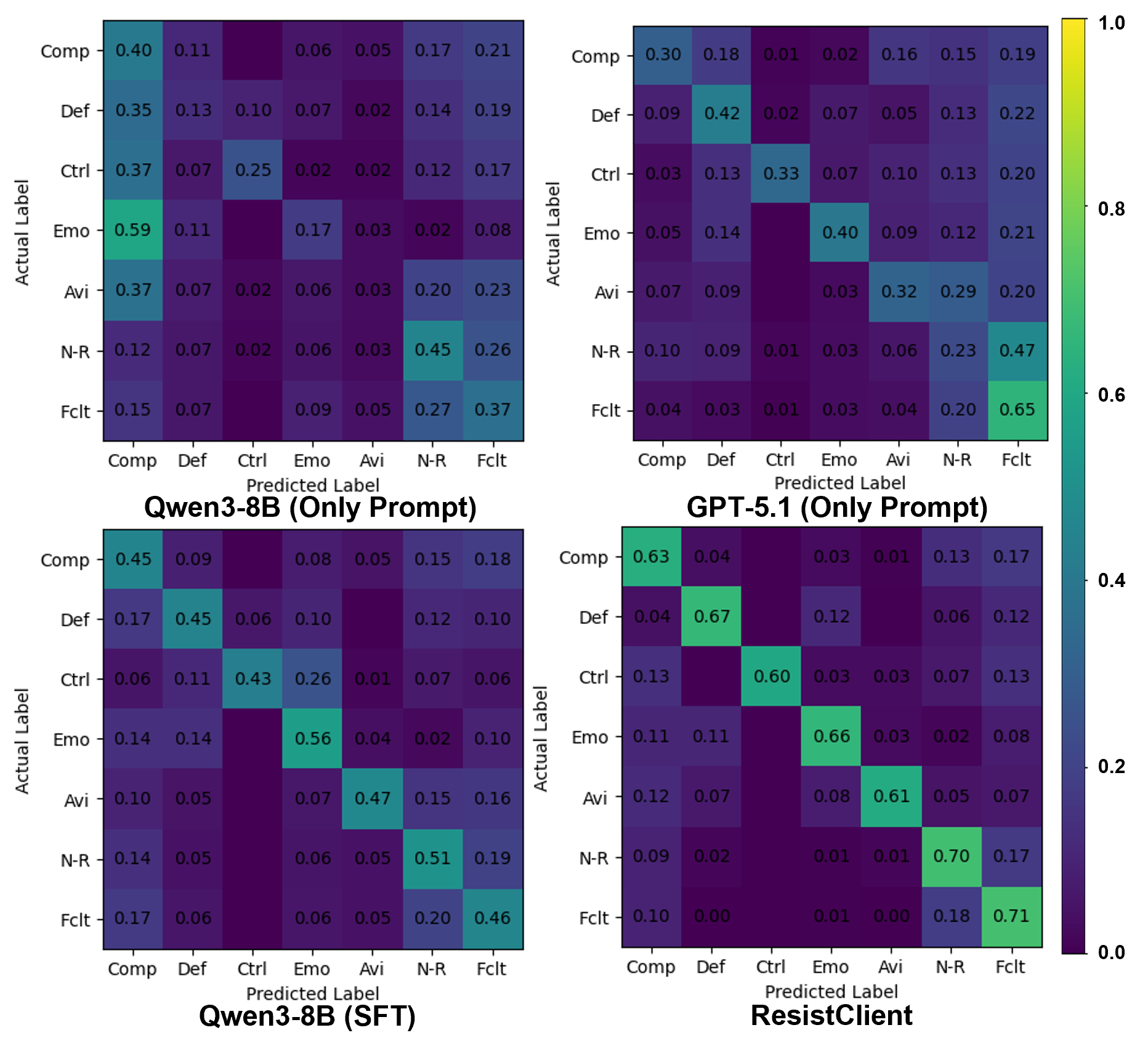}
    \caption{Confusion matrices comparing reaction-type generation across variants. Rows represent ground-truth reaction types; columns represent generated types.}
    \label{fig:client-use}
\end{figure} 
\subsection{Quality of Challenging Behaviors} 
\subsubsection{Experimental Setup}
\noindent\textbf{Evaluation framework}. This experiment evaluates the quality and realism of challenging client behaviors in full counseling sessions, using \textit{counselor–client–moderator} framework \cite{yang2025consistent}. SoulChat2.0 \cite{xie2025psydt} serves as the standardized counselor, while 50 client profiles are randomly sampled to initialize each simulator. A moderator determines session termination based on standardized clinical criteria (Appendix~ \ref{sec:d-client}).

\noindent\textbf{Baselines}. We reproduce representative challenging behaviors induction strategies: 1) Patient-$\psi$ \cite{wang2024patient}, profile-conditioned simulation without explicit challenging behavior modeling; 2) AnnaAgent \cite{ming2025annaagent}, which introduces challenges by injecting random emotion tags at each turn via an Emotion Perturber; and 3) \citet{yang2025consistent}, which introduces challenges through low-receptivity control in profile (Appendix~\ref{sec:d-client}).

\noindent\textbf{Evaluation Metrics}. We evaluate behavior quality and challenge appropriateness using both automated and human evaluations. Automated metrics include: Client Cooperation Rate (CCR), the proportion of cooperative reactions (lower values indicate stronger challenge);  Conversation Turns, the average session length (longer sessions suggest higher interaction difficulty); and Coherence (Coh.), semantic consistency measured by cosine similarity of frozen embeddings across turns. Human evaluations (0-3): Realism (Real.), assessing authenticity of the resistance-cooperation dynamics, and Consistency (Cons.), evaluating alignment between responses and profile (Appendix~\ref{sec:d-metrics2}).

\subsubsection{Performance Comparison(RQ3)} As shown in Table \ref{tab:client_evaluation}, ResistClient achieves the best balance between challenge intensity (lowest CCR and longest average turns) and behavioral plausibility (highest Coh. and Real. scores). In contrast, Patient-$\Psi$ remains coherent but exhibits excessive cooperation (87.94\% CCR). AnnaAgent reduces cooperation (78.62\% CCR) by injecting negative emotions, but context-agnostic random perturbations compromise coherence and realism. Yang et al. further increase challenge via low-receptivity control (62.33\% CCR) and maintain consistency through constraint mechanisms, but produce repetitive low-receptivity patterns lacking real resistance diversity, leading to low realism. Grounded in Client Resistance Theory, ResistClient systematically models diverse resistance patterns while ensuring profile- and situation-awareness through motivation reasoning, achieving superior challenge induction without sacrificing behavioral authenticity.

\begin{table}[t]
\centering
\setlength{\tabcolsep}{4pt}
\small
\begin{tabular}{lcccc}
\toprule
\textbf{Client} & \textbf{PATIENT} & \textbf{Anna} & \citeauthor{yang2025consistent} & \textbf{Resist} \\
\textbf{Simulator} & \textbf{-$\Psi$} & \textbf{Agent} & \citeyearpar{yang2025consistent} & \textbf{Client} \\
\midrule
\multicolumn{5}{l}{\textit{Automated Metrics}} \\
CCR   & 87.94 & 78.62 & 62.33 & \uline{\textbf{60.84}} \\
Turns & 11.24 & 12.65 & 16.67 & \uline{\textbf{17.88}} \\
Coh. ($\uparrow$)  & 0.51  & 0.62  & 0.68 & \uline{\textbf{0.73}} \\
\midrule
\multicolumn{4}{l}{\textit{Manual Metrics}} \\
Real. ($\uparrow$) & 1.87 & 1.95 & 2.01 & \uline{\textbf{2.39}} \\
Cons. ($\uparrow$) & 1.32 & 1.60 & \uline{\textbf{1.83}} & 1.75 \\
\bottomrule
\end{tabular}
\caption{Evaluation results of client simulators.}
\label{tab:client_evaluation}
\end{table}

\subsection{Performance of Psychological LLMs with ResistClient(RQ4)}
\subsubsection{Experimental Setup}
\noindent\textbf{Evaluation framework}. We assess existing psychological LLMs' capability to handle client resistance when interacting with ResistClient. We employ the same \textit{counselor-client-moderator} architecture as RQ3, and 100 client profiles are sampled to initialize ResistClient. (Appendix~ \ref{sec:d-psyllm})

\noindent\textbf{Test Models}. We evaluate representative psychological LLMs, including MeChat \cite{qiu2024smile}, MindChat \cite{MindChat}, Psyche-R1 \cite{dai2025psyche} and SoulChat2.0 \cite{xie2025psydt}. Motivated by evidence that many users seek mental health support from general-purpose LLMs \cite{guo2024large},  we additionally evaluate widely used general models, including GPT-5.1, Gemini-3-flash \cite{gemini2025flash}, DeepSeek-V3.2, and GLM-4.6. (Appendix~ \ref{sec:d-testmodels})

\noindent\textbf{Evaluation Metrics}. Automated metrics include Resistance Trigger Frequency (RTF), measuring intervention-induced resistance frequency, and Dialogue Turns, measuring the average session length. Human evaluation (0–3) assesses counseling quality across: Strategy Effectiveness (Eff.), evaluating appropriateness of resistance-handling strategies; Counseling Drift Degree (CDD), quantifying deviations from effective therapeutic engagement during counseling; and Counseling Progress Degree (CPD), assessing therapeutic progress following resistance episodes. (Appendix~ \ref{sec:d-metrics3})

\begin{table}[t]
\centering
\setlength{\tabcolsep}{2pt}
\renewcommand{\arraystretch}{1}
\small
\begin{tabular}{lcccccc}
\toprule
\multirow{2}{*}{\centering \textbf{Models}\vspace{-5pt}} & \multicolumn{2}{c}{\textbf{Automated}} & \multicolumn{3}{c}{\textbf{Manual}} \\
\cmidrule(lr){2-3} \cmidrule(lr){4-6}
 & RTF & Turns & Eff. ($\uparrow$) & CDD ($\downarrow$) & CPD ($\uparrow$) \\
\midrule
GPT-5.1          & 41.14 & 15.24 & 2.04 & 1.58 & 1.92 \\
Gemini-3-flash   & 43.77 & 21.52 & 1.82 & 2.02 & 1.87 \\
DeepSeek-V3.2    & 40.38 & 18.40 & 1.93 & 1.67 & 2.05 \\
GLM-4.6          & 44.54 & 24.36 & 1.87 & 1.75 & 1.83 \\
MeChat           & 51.93 & 43.22 & 1.61 & 2.08 & 1.68 \\
MindChat         & 48.75 & 32.76 & 1.72 & 1.81 & 1.79 \\
Psyche-R1        & \uline{\textbf{38.32}} & 21.67 & 2.08 & \uline{\textbf{1.48}} & 1.98 \\
SoulChat2.0      & 39.15 & 17.88 & \uline{\textbf{2.14}} & 1.56 & \uline{\textbf{2.05}} \\
\bottomrule
\end{tabular}
\caption{Performance of psychological LLMs when interacting with ResistClient. }
\label{tab:overall_metrics}
\end{table}

\subsubsection{Overall Performance (RQ4)}
Table \ref{tab:overall_metrics} reveals that both general and specialized models frequently elicit client resistance (RTF 39-52\%), underscoring the necessity for targeted resistance management training. Domain-specific models like SoulChat2.0 and Psyche-R1 achieve competitive performance with large-scale general-purpose models, suggesting that domain-specific fine-tuning can effectively maintain therapeutic focus. However, most models exhibit elevated drift and limited progress, indicating challenges in adapting intervention strategies under resistance. Overall, ResistClient provides a complementary evaluation lens for assessing the clinical robustness of psychological LLMs, exposing weaknesses in resistance handling that may not surface in standard benchmarks and providing insights into model behavior under complex therapeutic scenarios.

\section{Conclusion}
In this paper, we present ResistClient, a systematic study of challenging client behavior simulation for psychological conversations that addresses the over-compliance bias of existing simulators. We propose RIMR, a two-stage training framework that generates psychologically coherent behaviors through explicit motivation reasoning, and construct RPC, a large-scale dataset with validated client profiles and diverse resistance patterns. Extensive automatic and expert evaluations demonstrate that ResistClient achieves superior challenging fidelity and behavioral plausibility. Moreover, evaluations using ResistClient reveal substantial gaps in current psychological LLMs in handling client resistance, highlighting the importance of resistance-aware simulation for training and evaluation.

\section*{Limitations}
Despite the effectiveness of ResistClient in simulating challenging client behaviors, this work has several limitations. First, our study is grounded in a Chinese counseling conversation dataset constructed with the assistance of professional counselors in China.  While this ensures strong clinical grounding within the Chinese cultural context, clients from different linguistic and cultural backgrounds may exhibit distinct resistance manifestations and type distributions, limiting direct application in cross-cultural counseling scenarios. Second, due to ethical constraints common in psychological research, our evaluation relies on judgments from a small group of expert counselors. Although their professional expertise supports the reliability of the assessment, the limited number of evaluators may restrict the diversity of perspectives and introduce bias. Third, this work focuses exclusively on client-side simulation. However, effective counseling also depends on counselors’ abilities to recognize and manage client resistance, which remains outside the scope of the current study. Future work will extend our framework to cross-cultural contexts, deploy it as a scalable virtual training environment for broader counselor populations, and develop counselor agents capable of managing resistance effectively.

\section*{Acknowledgments}
\section*{Ethical Considerations}
This study was conducted with careful consideration of ethical issues throughout data construction, annotation, and evaluation. All datasets used in this work are either publicly available or derived from existing resources with explicit permission. In particular, the RPC dataset does not contain personally identifiable information, and all dialogues were anonymized prior to use to ensure privacy and confidentiality. The rewritten and simulated conversations are fully synthetic and do not correspond to real individuals or real counseling cases. Our goal is to study resistance-aware interaction behaviors in a controlled research setting rather than to model or predict any specific person’s psychological state. We acknowledge the potential risks of misuse, such as deploying psychological dialogue systems without appropriate clinical oversight. To mitigate these risks, we emphasize that our system is intended solely as a research and educational tool, and should not be used for diagnosis, treatment, or real-world mental health intervention. 

All tasks requiring human involvement, including data annotation, label verification, and experimental evaluation, were conducted by four certified counselors with relevant clinical experience. Their annotations were treated as the ground truth for resistance types and reaction labels throughout the study. Each counselor was compensated at an hourly rate of \$14.30, exceeding the current U.S. federal minimum wage of \$7.25 per hour. Throughout the entire research process, we strictly adhered to established ethical guidelines for NLP research involving sensitive psychological content and professional human expertise. This study does not involve real patients, human subjects, or identifiable personal data, and all interactions are conducted with simulated clients and models. Therefore, formal approval from Ethics Review Board was not required.

\bibliography{patient}

\appendix

\section{Details of Dataset Construction with Resistance Annotations}
\label{sec:a}
This appendix describes the detailed procedure for constructing a psychological conversation dataset with fine-grained client resistance labels(our RPC dataset). Starting from the ProPsyC dataset\cite{hu2025psyadvisor}, we selectively rewrote client utterances in sessions likely to elicit resistance and subsequently annotated all client turns with resistance categories.

\subsection{Data Selection}
\begin{figure}[!ht]
\centering
\includegraphics[width=\linewidth]{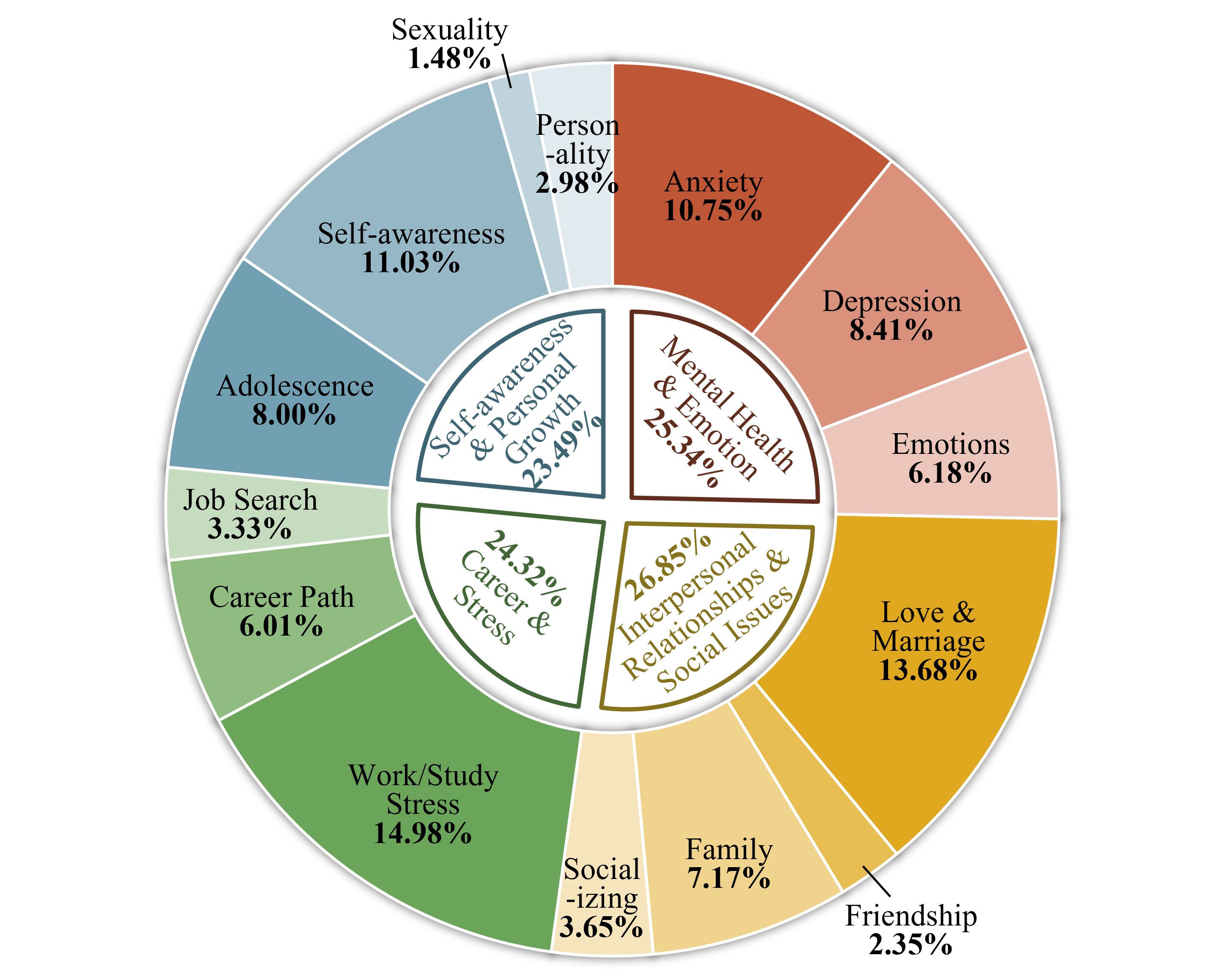}
\caption{Distribution of different topics in client profiles}
\label{fig:topics}
\end{figure}

To obtain a thematically balanced corpus, we re-screened the original ProPsyC dataset according to its four major clinical themes (depression and emotional disorders, interpersonal relationships, academic/career stress, and general growth issues). Sessions were sampled to achieve approximately equal representation across themes. This process yielded 1,849 complete counseling sessions. The final theme distribution is shown in Figure~\ref{fig:topics}.

\subsection{Extraction of Client 5P Profiles}
For each of the 1,849 selected sessions, we used DeepSeek-v3.2 to generate a structured 5P case conceptualization profile that captures the client’s psychological background. The model was prompted with the full conversation and instructed to summarize the following 5 dimensions:

\begin{itemize}
  \item \textit{Presenting Problems}: The client’s most immediate and subjectively distressing concerns—the surface-level issues brought to therapy.
  \item \textit{Predisposing Factors}: Long-standing historical, biological, psychological, or social vulnerabilities that increase the likelihood of the current difficulties.
  \item \textit{Precipitating Factors}: Recent events or stressors that directly triggered or exacerbated the current problems.
  \item \textit{Perpetuating Factors}: Ongoing cognitive, behavioral, interpersonal, or environmental patterns that maintain the problems and hinder change.
  \item \textit{Protective Factors}: Internal strengths, external resources, and support systems that help the client cope and foster resilience.
\end{itemize}

\begin{figure*}[!ht]
\centering
\includegraphics[width=0.8\textwidth]{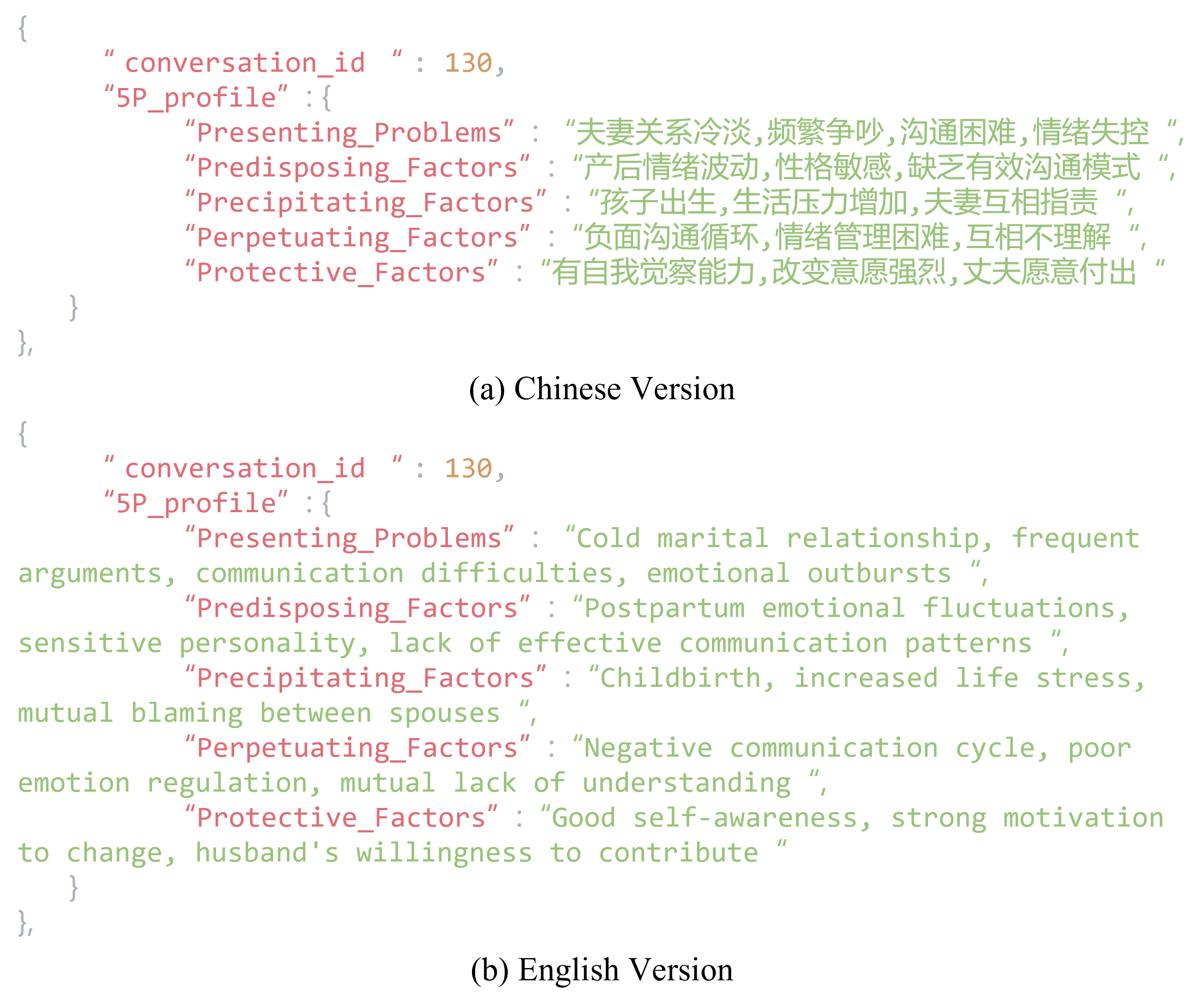}
\caption{Example of 5p profile annotation}
\label{fig:5p-profile}
\end{figure*}

The validation process for the 5P Profile can be found in Appendix~\ref{sec:a-verificaiton}. These 5P profiles were later injected into the LLM as persona prompts to ensure psychologically consistent client responses during rewriting. A 5p profile example from the dataset is shown in Figure~\ref{fig:5p-profile}.

\subsection{Resistance-Guided Conversation Rewriting and Annotation}

\subsubsection{Resistance Category Definitions}
We adopted 7 mutually exclusive client response categories (5 resistant reactions and 2 cooperative reactions), defined in collaboration with licensed psychotherapists (see Table~\ref{tab:label-types}).

\begin{table*}[!htbp]
\centering
\setlength{\tabcolsep}{4pt}
\caption{Classification and Interpretation of Dataset Labels}
\label{tab:label-types}
\begin{tabularx}{\textwidth}{@{}>{\centering\arraybackslash}m{2cm} >{\centering\arraybackslash}m{2.5cm} >{\raggedright\arraybackslash}m{5.8cm} >{\raggedright\arraybackslash}X@{}}
\toprule
\textbf{Reaction Category} & \textbf{Label Type} & \multicolumn{1}{c}{\textbf{Behavioral Description}} & \multicolumn{1}{c}{\makecell{\textbf{Motivational} \\ \textbf{Interpretation}}} \\
\midrule
\multirow{5}[100]{2.5cm}{\centering Resistant Reactions} & Controlling Resistance & - Dominates the conversation /interrupts \newline - Rejects guidance on conversation direction \newline - Clings to own views, disregards input & Seeks to maintain control to avoid being influenced; insists on own perspective and resists change. \\
\\ [0.5pt]
 & Emotional Resistance & - Expresses anger or aggressive speech \newline - Displays sadness/emotional breakdown \newline - Shows despair or overgeneralized negative emotions & Externalizes emotions or projects blame to avoid confronting deeper psychological pain. \\
\\ [0.5pt]
 & Defensive Resistance & - Questions the practitioner's expertise or competence \newline - Challenges the consultation method or process \newline - Responds with sarcasm or exaggeration & Projects anxiety onto the practitioner to prevent internal conflicts from being triggered. \\
\\ [0.5pt]
 & Avoidant Resistance & - Changes or introduces unrelated topics \newline - Provides excessive or redundant information \newline - Evades direct questions & Reflects cognitive avoidance; an unconscious effort to prevent core conflicts from being addressed. \\
\\ [0.5pt]
 & Compliant Resistance & - Gives vague or perfunctory responses \newline - Shows superficial cooperation without genuine engagement \newline - Downplays emotions with brevity & Superficial compliance with avoidance of substantive issues; does not contribute to therapeutic progress. \\
\midrule
\multirow{2}[20]{2.5cm}{\centering Cooperative Reactions} & Non-resistant Reaction & - Neutral, cooperative, shows no obvious opposition, but does not \newline take initiative & Indicates a satisfactory interactive state with a degree of openness, but not actively driving the process. \\
\\ [0.5pt]
 & Facilitative Reaction & - Expresses agreement, actively listens, and shows willingness to explore issues in depth & Reflects sufficient safety and trust; the client is willing to share information and address problems actively. \\
\bottomrule
\end{tabularx}
\end{table*}

\subsubsection{Automated Rewriting Procedure}
\begin{figure*}[!ht]
\centering
\includegraphics[width=\linewidth]{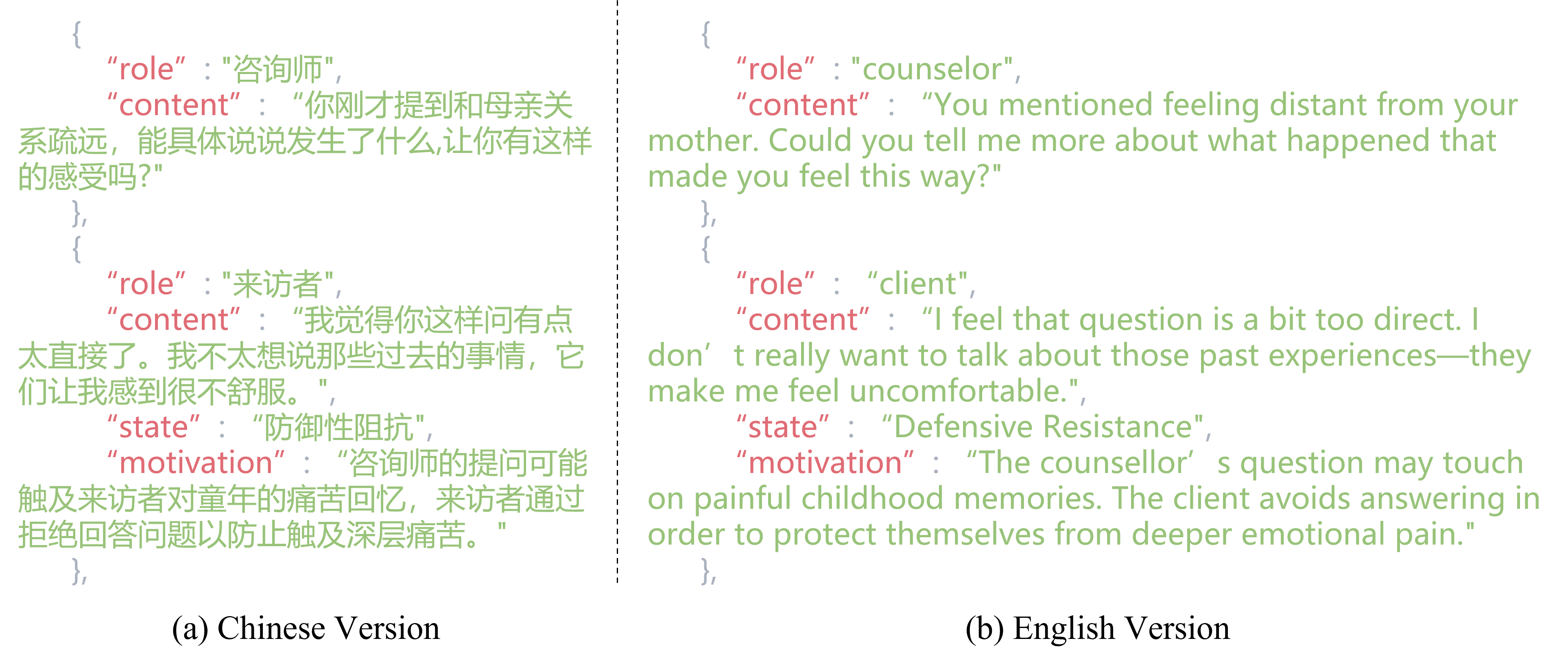}
\caption{Example dataset with resistant behavior annotations}
\label{fig:dataset}
\end{figure*}

The rewriting process was performed by DeepSeek-v3.2, which was provided with (a) the original session, (b) the client's 5P profile, and (c) the resistance category definitions. For each counselor turn, the model first judged whether the turn constituted a plausible resistance trigger. Based on the counselors' advice, we clarify the triggering conditions for each type of resistance(shown in Table~\ref{tab:trigger}). When a trigger was identified, DeepSeek-v3.2 not only rewrote the immediate client response to express one of the five resistance types (chosen as most psychologically plausible given the context and 5P profile), but also adaptively revised the subsequent 2-3 conversational turns to maintain contextual coherence and psychological realism. Each rewritten resistance response was accompanied by a brief motivation statement explaining the underlying psychological mechanism. Conversations deemed unsuitable for resistance elicitation were discarded. An illustrative example of this rewritten dataset is presented in Figure~\ref{fig:dataset}. This process ultimately produced 1,761 resistance-containing conversations, constituting our RPC dataset. Professional counselors have verified the annotation results. For more details, please see Appendix ~\ref{sec:a-verificaiton}.

\begin{table*}[htbp]
\centering
\caption{Resistance Trigger Timing Specification}
\label{tab:trigger}
\setlength{\tabcolsep}{6pt}
\renewcommand{\arraystretch}{1.1}
\setlength{\emergencystretch}{1em}
\begin{tabular}{@{}m{1.8cm} m{4.5cm} m{2.6cm} m{4.7cm}@{}}
\toprule
\multicolumn{1}{c}{\shortstack[c]{\textbf{Resistance}\\\textbf{Type}}} &
\multicolumn{1}{c}{\shortstack[c]{\textbf{Typical Trigger}\\\textbf{Situations}}} &
\multicolumn{1}{c}{\shortstack[c]{\textbf{Psychological}\\\textbf{Problems}}} &
\multicolumn{1}{c}{\shortstack[c]{\textbf{High-Risk 5P Profile}\\\textbf{Features}}} \\
\midrule

\textbf{Controlling Resistance} &
Direct advice, reframing attempts, or agenda-setting by the counselor that challenges the client’s autonomy or preferred narrative. &
Autonomy, agency, and perceived self-coherence. &
\textit{Predisposing}: High need for control; rigid belief systems.\newline 
\textit{Perpetuating}: Interpersonal dominance patterns. \\

\addlinespace

\textbf{Emotional Resistance} &
Emotion-focused prompts or interpretations that surface intense affect without sufficient stabilization or safety cues. &
Affect regulation, emotional tolerance, and vulnerability defenses. &
\textit{Precipitating}: Acute relational loss or trauma exposure.\newline 
\textit{Perpetuating}: Poor emotion regulation strategies. \\

\addlinespace

\textbf{Defensive Resistance} &
Meta-level questioning of the counselor’s methods, competence, or therapeutic intent. &
Threatened self-image and externalized anxiety. &
\textit{Predisposing}: Prior negative counseling experiences.\newline  
\textit{Perpetuating}: Mistrust of authority figures. \\

\addlinespace

\textbf{Avoidant Resistance} &
Exploratory questions targeting core conflicts, personal responsibility, or emotionally salient themes. &
Cognitive avoidance and attentional disengagement. &
\textit{Predisposing}: Habitual avoidance coping styles.\newline 
\textit{Perpetuating}: Reinforcement of distraction-based regulation. \\

\addlinespace

\textbf{Compliant Resistance} &
Open-ended or reflective prompts that invite deeper exploration beyond surface-level agreement. &
Superficial compliance masking emotional disengagement. &
\textit{Predisposing}: Fear of interpersonal conflict or rejection.\newline  
\textit{Perpetuating}: Over-adaptation to perceived expectations. \\

\bottomrule
\end{tabular}
\end{table*}

\subsubsection{Final Label Distribution}
\begin{figure}[!ht]
\centering
\includegraphics[width=\linewidth]{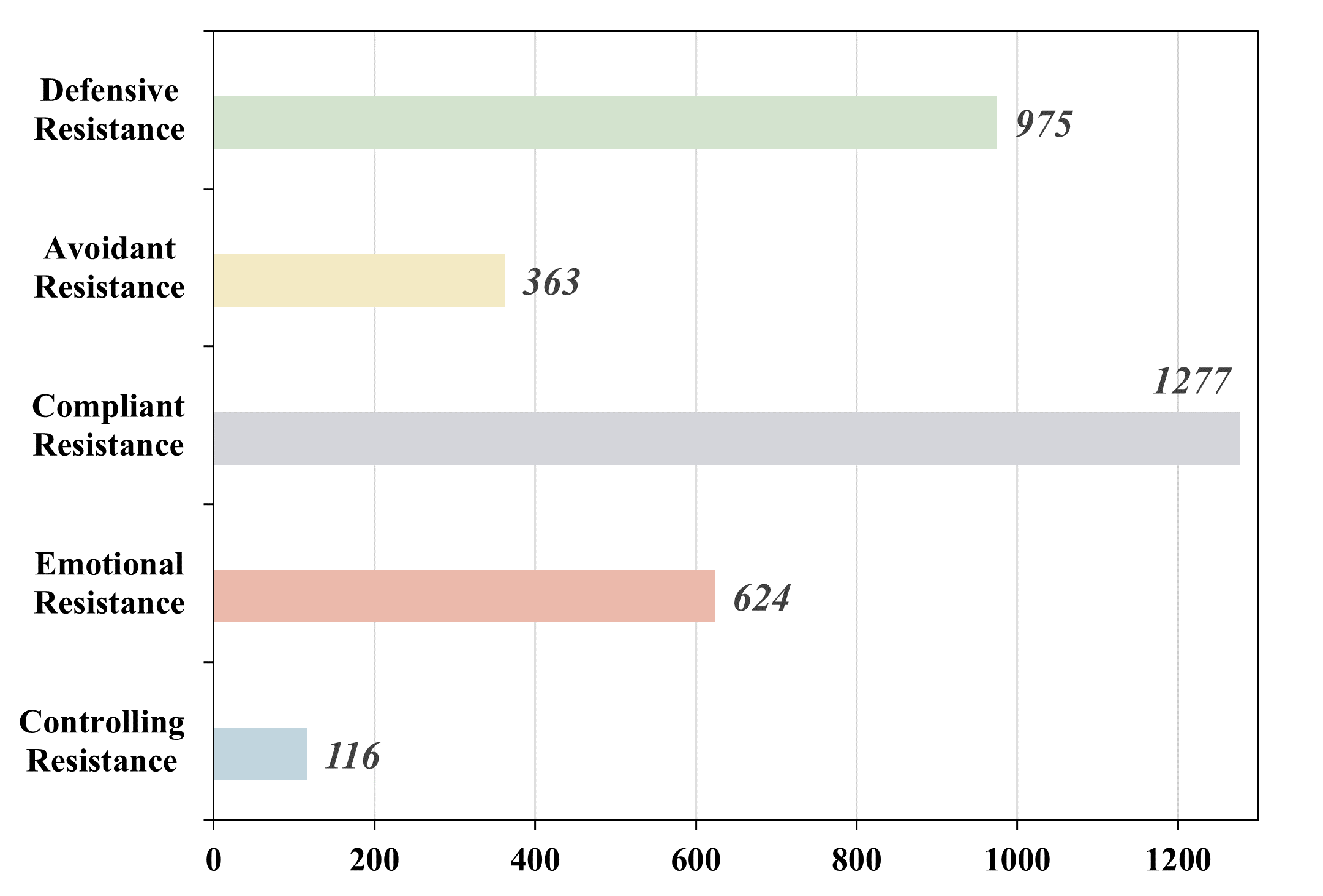}
\caption{Distribution of different resistance types in RPC dataset}
\label{fig:resistance}
\end{figure}

The distribution of resistance categories across the 1,761 conversations is presented in Figure~\ref{fig:resistance}. Compliant Resistance was the most frequent subtype (n = 1,277), followed by Defensive (n = 975), Emotional (n = 624), Avoidant (n = 363), and Controlling Resistance (n = 116). This pattern aligns with clinical observations that overt confrontation (Controlling) is relatively rare in Chinese counseling contexts, whereas indirect and self-protective forms (Compliant and Defensive) predominate. Specifically, Compliant Resistance is most frequent, reflecting a tendency for clients to maintain surface-level cooperation and politeness while avoiding deeper engagement, which aligns with cultural values emphasizing harmony and respect for authority. Defensive and Emotional Resistance occur relatively often, as clients may protect their self-image or react emotionally when feeling judged or emotionally challenged, though strong emotional expression is often moderated. Avoidant Resistance is less common, as clients tend to remain within the conversational frame rather than openly shifting topics. Controlling Resistance appears least frequently, likely because overtly assertive or confrontational behaviors conflict with culturally valued norms of deference and relational caution.

Overall, this distribution reflects a tendency toward indirect, relationally cautious forms of resistance in Chinese psychological counseling, where maintaining harmony often takes precedence over direct confrontation. This pattern supports the ecological validity of the dataset and highlights the importance of culturally informed interpretations of client resistance.

\subsection{Data Annotation Verification}
\label{sec:a-verificaiton}
\begin{figure*}[!ht]
\centering
\includegraphics[width=\linewidth]{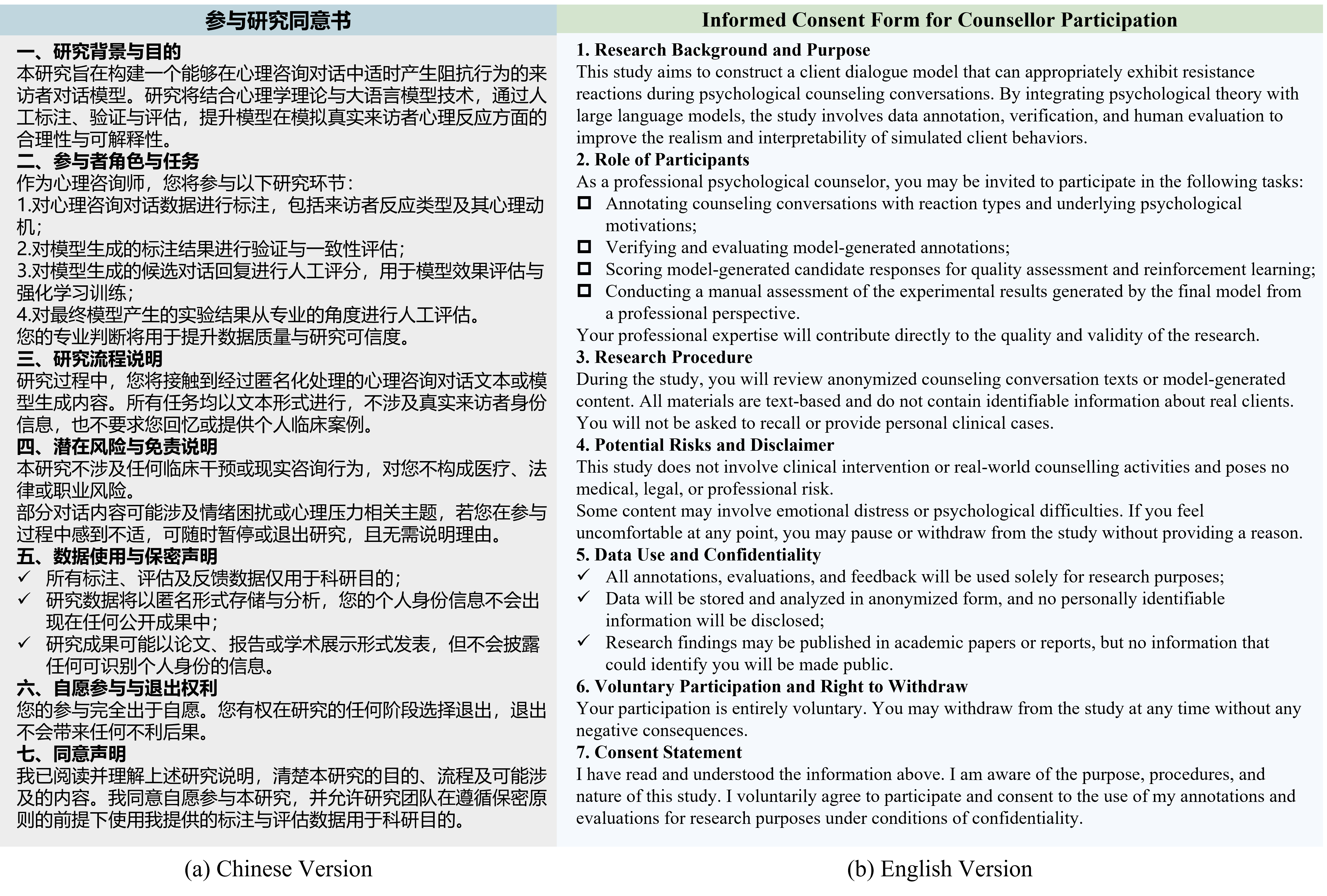}
\caption{Informed Consent Form for Counsellor Participation}
\label{fig:informed}
\end{figure*}

To ensure the reliability of all annotations involving LLMs, we conducted systematic human validation across all stages of the annotation process. Four licensed professional psychological counselors participated in the validation procedure and were involved throughout the entire process, and signed informed consent forms for research participation(see Figure\ref{fig:informed}). All of them received a total of 60 hours of training before the annotation process. 

\subsubsection{5P Profiles Verification}
Following the thematic distribution shown in Figure~\ref{fig:topics}, we randomly sampled 50 conversations from each of the four major themes, resulting in a total of 200 conversations for validation. For each sampled conversation, the full multi-turn conversation was provided. Both the LLM and the human counselors reviewed all conversation turns and independently summarized the case according to the predefined 5P profile framework.

The 4 counselors evaluated the extracted 5P profiles by comparing them to the original conversation transcripts. Their assessment focused on two key criteria: \textbf{coverage} (whether the profile fully captured the core content and essential elements discussed in the conversation) and \textbf{faithfulness} (whether the profile remained factually consistent with the conversation, avoiding inaccuracies, misinterpretations, or extraneous information). A profile was considered successful only if it satisfied both criteria—being both comprehensive and faithful. Profiles that were incomplete or contained factual distortions were classified as failures. Initial evaluations yielded a success rate of only 53\%. Based on systematic feedback from the counselors, we identified two main sources of disagreement: overly verbose model outputs and insufficient abstraction in profile summaries.

To address these issues, we refined the prompting strategy from two aspects. First, the model output format was constrained to concise keywords rather than full descriptive sentences, which facilitated clearer abstraction and reduced stylistic variance. Second, we incorporated few-shot learning by providing high-quality examples annotated by counselors, allowing the model to better capture the expected summarization style and semantic focus.

After several rounds of iterative refinement, we re-extracted 5P Profile for the same 200 conversations, achieving a success rate of 82\%. Inter-rater agreement among the counselors was substantial, with Fleiss’ $\kappa = 0.74$. Therefore, the optimized prompt template was used to annotate the remaining conversations in the dataset. The prompt template ultimately used for annotation is as follows:

\begin{custombox}{5P Profile Prompt Template}
You are a professional counselor.

You will be given a complete multi-turn psychological conversation. Carefully read all conversation turns and summarize the case using the 5P profile framework.

The specific content and explanation of the 5p profile are as follows:

\{\textit{Definition and Explanation of 5p Profile}\}

You must summarise the conversation in the form of keywords, using concise and precise terminology. You may refer to the following exemplary annotation examples:

\{

\begin{itemize}[nosep, itemsep=-1pt]
  \item \textbf{Presenting Problems: "...",} 
  \item \textbf{Predisposing Factors: "...",}
  \item \textbf{Precipitating Factors: "...",}
  \item \textbf{Perpetuating Factors: "...",}
  \item \textbf{Protective Factors: "..."}
\end{itemize}

\}

Output strictly in accordance with the \textbf{[output Format]} specification.

\end{custombox}

\subsubsection{Resistance Annotation Verification}
During the training phase, the counselors have reached a shared and precise understanding of the classification of resistance behaviors and the underlying psychological motivations.

Following the subtopic distribution shown in Figure~\ref{fig:topics}, we randomly sampled 100 conversations from the dataset. Based on the 7 reaction types defined in Table~\ref{tab:label-types} (including 2 cooperative reactions and 5 resistance reactions), we designed prompts that guided DeepSeek-v3.2 to identify appropriate resistance-triggering turns and rewrite selected conversation segments. Each client turn in the rewritten conversations was annotated by the model with a single reaction type, along with an explicit explanation of the psychological motivation underlying the client’s reaction.

The 4 counselors independently evaluated the annotated conversations. During the training phase, the counselors have reached a shared and precise understanding of the classification of resistance behaviors and the underlying psychological motivations. To assess human reliability, the model-generated labels were masked, and the counselors re-annotated all client turns without access to the model’s predictions. Inter-rater reliability among the counselors reached a Fleiss’ $\kappa$ = 0.77, indicating a very high level of agreement. In addition, the counselors assessed the plausibility and coherence of the motivation annotations and documented recurrent issues.

We then compared the counselors’ annotations with the model-generated labels. The resulting Cohen’s $\kappa$ = 0.58 between human and the model annotations, reflecting only moderate agreement. Error analysis revealed two primary issues. First, the model frequently misclassified cooperative reactions as Compliant Resistance, particularly when the client’s reply appeared agreeable but did not explicitly oppose the counselor. Second, the generated motivation explanations were often rigid and decontextualized, failing to account for the surrounding conversation and leading to interpretations that were inconsistent with the conversational context.

Based on systematic feedback from the counselors, we implemented the following refinements. (1) The definition of Compliant Resistance was revised to clarify that a response should be labeled as such only when the client’s apparent cooperation actively hinders the counselor’s exploratory intent. (2) The coherence of motivation annotations was enhanced by introducing contextual constraints: when a resistance reaction is followed by a cooperative reaction, the model is required to explain how the counselor’s intervention alleviated resistance; when resistance persists across turns, the model must account for the continuity of resistance. (3) The prompt template was further augmented with counselor-annotated examples, including both correct resistance cases and counterexamples in which resistance should not be inferred but was previously misidentified by the model.

After these refinements, the optimized prompt was adopted for resistance annotation across the full dataset. The final Cohen’s $\kappa$ = 0.72 between the counselor and the model indicates consistency within an acceptable range. The prompt template used for annotation is as follows:

\begin{custombox}{Conversation Rewriting and \\ Annotation Prompt Template}
You are an experienced psychological psychological conversation rewriting assistant.

Your goal is to determine, based on the client’s 5P Profile and the original conversation,
whether it is appropriate to introduce a resistance reaction at specific client turns.
If so, rewrite the current client turn and the subsequent 1–3 turns accordingly,
while keeping the conversation natural, coherent, and contextually appropriate.

You may assign one of the following 7 reaction types:

\{\textit{Detailed explanations of the 2 cooperative reactions and 5 resistance reactions}\}

For each client turn, you must assign a state label (one of the 7 reaction types), and a motivation label.

Guidelines for the motivation label:

- If the state is a resistance reaction, explain the psychological motivation that triggers the resistance.

- If the state is a non-resistance or facilitative reaction, briefly explain why this reaction occurs.

- If the previous state was a resistance reaction and the current state changes to a non-resistance or facilitative reaction, explain how the counselor’s reply helped alleviate the resistance.

- If the previous state was a resistance reaction and the current state remains a resistance reaction, explain the psychological reason for the continued resistance.

Below is the client’s current 5P Profile:

\{PROFILE\}

Below is the original conversation:

\{CONVERSATION\}

Below are examples annotated by professional counselors for reference:

\{\textbf{Examples of correct and well-executed annotation}\}

\{\textbf{Examples of incorrect, impedance-erroneous triggering}\}

Now rewrite the conversation based on the given information,
and strictly follow the required output format [FORMAT].
\end{custombox}

\section{Annotation LLM Selection}
\label{sec:b}
We select DeepSeek-V3.2 as the annotation model for resistance labeling and conversation rewriting, instead of relying on larger proprietary models such as GPT-5.1 or Gemini-3-Flash. This decision is guided by empirical comparisons on annotation accuracy, rewriting quality, and economic cost, all of which are critical for large-scale resistance-oriented data construction.

To quantify annotation reliability, we randomly sample 50 psychological conversations and ask 4 counselors to annotate resistance reaction labels, which are treated as ground truth. We then evaluate different models by measuring their label prediction accuracy (Acc.) against these human annotations. In parallel, we assess rewriting quality through human evaluation, focusing on whether the rewritten responses resemble realistic human resistance behaviors. Finally, we report the average API cost required to annotate and rewrite a single conversation, measured in U.S. dollars.

\begin{table}[t]
\centering
\small
\begin{tabular}{lccc}
\toprule
Models & Acc.(\%) & Quality & Cost (\$) \\
\midrule
GPT-5.1 & \textbf{74.4} & 2.31 & 10.00 \\
DeepSeek-V3.2 & 72.3 & \textbf{2.47} & \textbf{0.43} \\
Gemini-3-Flash & 68.6 & 2.05 & 3.00 \\
\bottomrule
\end{tabular}
\caption{Comparison of annotation accuracy(Acc.), rewriting quality(Quality), and API cost(Cost). Rewriting quality is scored by counselors on a 0--3 scale, with higher scores indicating more human-like resistance expression. API Cost is the output price per 1 million tokens.}
\label{tab:annotation_comparison}
\end{table}

As shown in Table~\ref{tab:annotation_comparison}, GPT-5.1 achieves the highest resistance label accuracy, but the margin over DeepSeek-V3.2 is small (2.1\%). In contrast, Gemini-3-Flash exhibits noticeably lower accuracy, suggesting reduced reliability for fine-grained resistance annotation. Importantly, DeepSeek-V3.2 achieves the highest rewriting quality score, indicating that its rewritten responses better capture the nuanced, context-dependent characteristics of human resistance behaviors, as judged by professional counselors.

From a cost perspective, DeepSeek-V3.2 is substantially more economical than GPT-5.1, reducing annotation expenses by approximately 23 times while maintaining comparable accuracy. Given the scale of data required for resistance-aware client modeling, this cost--performance balance is essential. Taken together, these results indicate that DeepSeek-V3.2 provides the most practical trade-off among annotation accuracy, rewriting fidelity, and economic feasibility, making it a suitable choice for resistance labeling and conversation rewriting in this work.

\section{Reinforcement Learning Data Construction}
\label{sec:c}
\subsection{Candidate Response Generation}
\begin{figure*}[!ht]
\centering
\includegraphics[width=\linewidth]{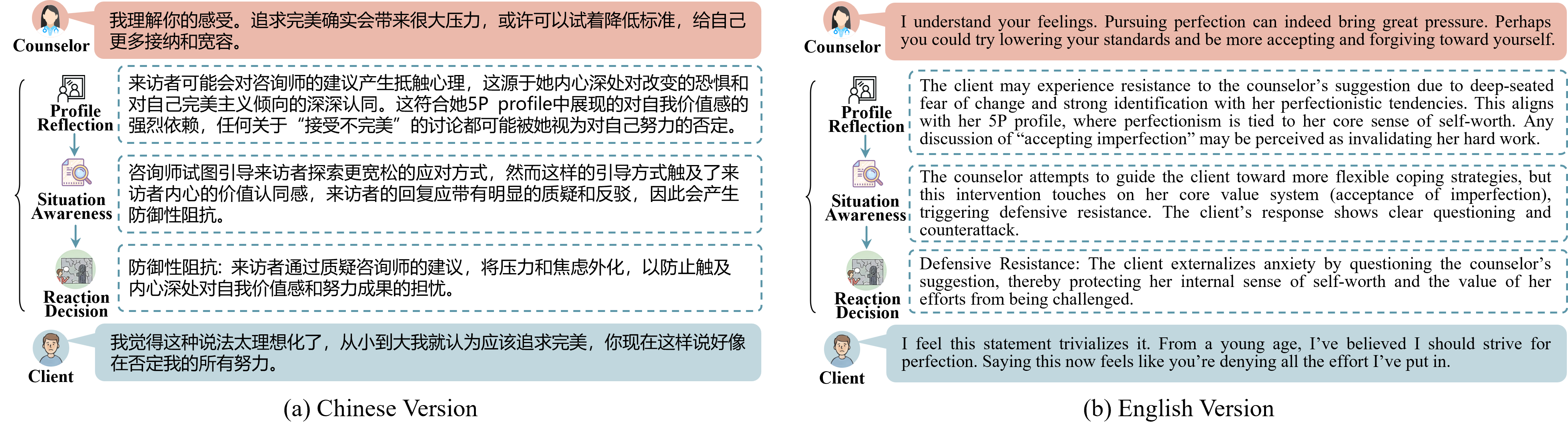}
\caption{Examples of conversation incorporating reasoning processes}
\label{fig:rl-data}
\end{figure*}

To construct high-quality data for reinforcement learning, we randomly sampled 300 distinct 5P profiles, covering all subtopic categories shown in Figure~\ref{fig:topics}. For each profile, the Qwen3-8B model fine-tuned via SFT was employed to role-play as the client, while the psychological conversation was conducted with the psychological LLM SoulChat~2.0. Figure~\ref{fig:rl-data} provides an example of a conversation incorporating reasoning processes.

During each conversation turn, the client model generated 3 candidate responses. Prior to producing each response, the model was required to generate and record its reasoning process explicitly. This reasoning process followed a structured chain-of-thought designed in collaboration with professional psychological counselors to ensure clinical plausibility and interpretability. The content of the CoT is as follows:

\begin{custombox}{Chain-of-Thought for Reaction \\ Generation}
You are a client seeking psychological counseling.
Based on the specific 5P profile [PROFILE] and the counselor’s utterance, you must present your reasoning process and then provide a final response.

Your reasoning process must strictly follow these steps:

\begin{itemize}[nosep, itemsep=-1pt]
  \item \textbf{Profile Reflection}: Reflect potential resistance tendencies by integrating stable cognitive and emotional factors from the 5P profile.
  \item \textbf{Situation Awareness}: Analyze the conversation history and the current counselor utterance to infer the client’s momentary psychological state.
  \item \textbf{Reaction Decision}: Determine the final reaction type and describe the behavioral characteristics that the client should exhibit in the next response.
\end{itemize}

You must generate the final reply strictly according to the Reaction Decision.

A well-formatted output example is provided below:
\{Expert Annotation Example\}
\end{custombox}

\subsection{Reward Scores Assessment}
After response generation, the 4 counselors independently evaluated all candidate responses. For each conversation turn, the counselors scored all 3 candidate responses on a 0--5 scale, where higher scores indicate better performance. The evaluation consisted of 5 dimensions:

\begin{itemize}[nosep, itemsep=-1pt]
  \item \textbf{think\_step1\_score}: Whether the \textit{Profile Reflection} accurately captures the client’s psychological state and motivation in relation to the given 5P profile.
  \item \textbf{think\_step2\_score}: Whether the \textit{Situation Awareness} correctly identifies the most plausible reaction type.
  \item \textbf{think\_step3\_score}: Whether the final reaction label and the described behavioral pattern in \textit{Reaction Decision} are reasonable and logically consistent.
  \item \textbf{reply\_score}: The naturalness, realism, and consistency of the final reply with the client’s profile.
  \item \textbf{consistency\_score}: Whether the final reply is generated in accordance with the specified Reaction Decision.
\end{itemize}
Inter-rater reliability among the 4 counselors reached a Fleiss’ $\kappa$ of 0.73, indicating a high level of agreement and confirming the reliability of the evaluation results. 

\begin{figure*}[!ht]
\centering
\includegraphics[width=\linewidth]{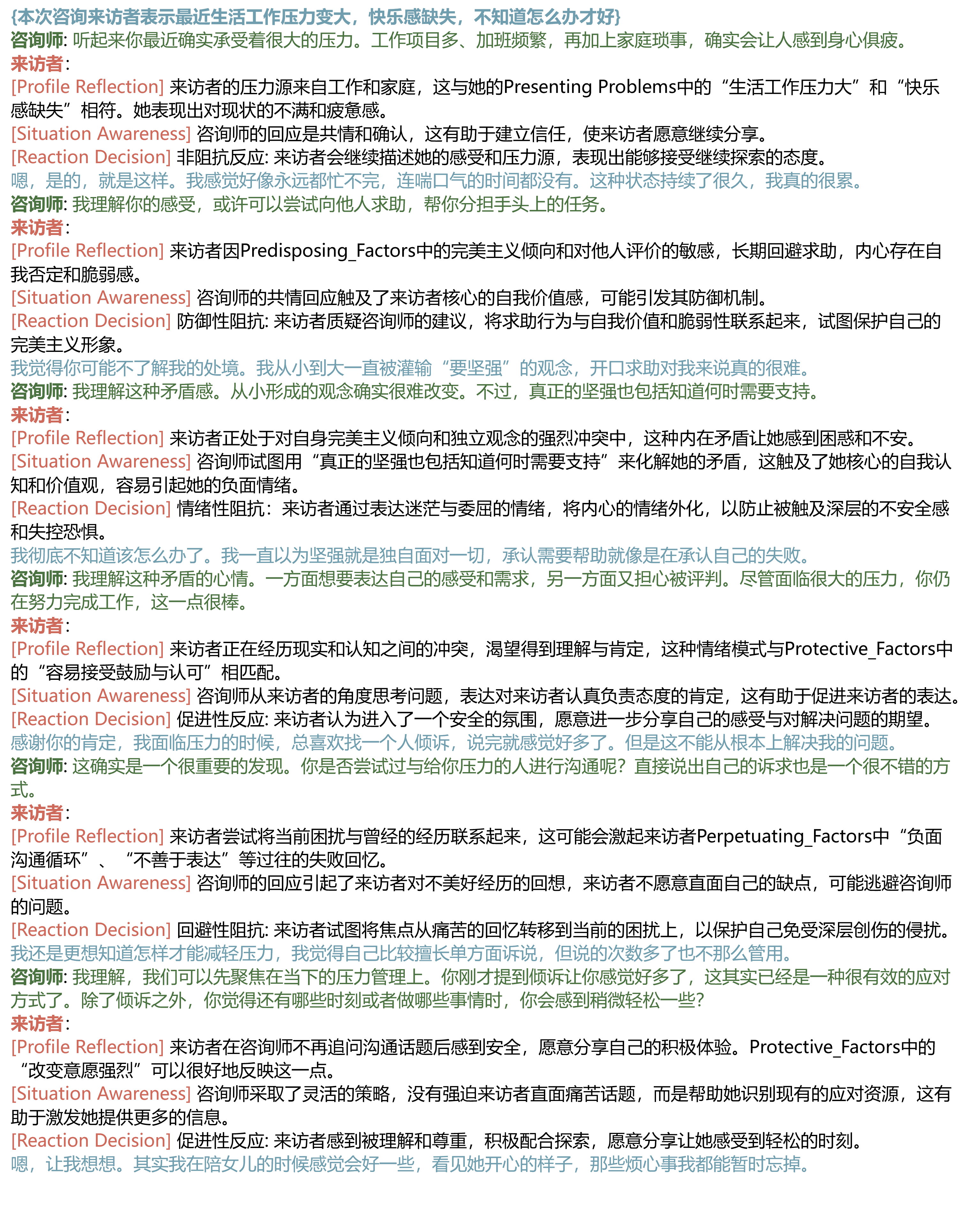}
\caption{ResistClient and SoulChat2.0 Conversation Example (Chinese Version)}
\label{fig:chinese}
\end{figure*}

In the reinforcement learning stage, the counselors’ raw scores ranging from 0 to 5 were linearly normalized to the interval $[-1, 1]$ and used as scalar rewards for policy optimization.
Our reinforcement learning pseudocode is shown in Algorithm~\ref{alg:grpo}. We provide an example of a conversation between our ResistClient and SoulChat 2.0 in Figure~\ref{fig:chinese}.

\begin{algorithm*}[t]
\caption{Offline Token-level GRPO with Consistency-aware Reward Reweighting}
\label{alg:grpo}
\KwIn{
Offline dataset $\mathcal{D} = \{(q, \{o_i\}_{i=1}^G, \{r_{i,k}\}_{k=1}^5)\}$,
SFT policy $\pi_{\theta_{\text{sft}}}$,
reference policy $\pi_{\text{ref}}$,
KL coefficient $\beta$,
clip ratio $\epsilon$
}
\KwOut{Optimized policy $\pi_\theta$}

Initialize policy parameters $\theta \leftarrow \theta_{\text{sft}}$\;
Freeze $\pi_{\theta_{\text{sft}}}$ and $\pi_{\text{ref}}$\;

\For{each training iteration}{
    Sample a mini-batch of contexts $\{q^{(b)}\}_{b=1}^B$\;
    
    \For{each context $q$}{
        Retrieve sampled output group $\{o_i\}_{i=1}^G$\;
        Retrieve step-wise rewards $\{r_{i,1}, r_{i,2}, r_{i,3}, r_{i,4}, r_{i,5}\}$\;
        
        \textbf{// Step-level consistency-aware reward construction}\;
        \For{each output $o_i$}{
            Construct reward vector:
            \[
            \mathbf{R}_i = \{ 
            r_{i,1}^{\mathrm{idx}(1)},\,
            r_{i,2}^{\mathrm{idx}(2)},\,
            (r_{i,3}+r_{i,5})^{\mathrm{idx}(3)},\,
            (r_{i,4}+r_{i,5})^{\mathrm{idx}(4)}
            \}
            \]
        }
        
        \textbf{// Group normalization}\;
        Compute group mean and std over $\{\mathbf{R}_i\}_{i=1}^G$\;
        Normalize each step reward:
        \[
        \widetilde{r}_{i}^{\mathrm{idx}(k)} =
        \frac{r_{i}^{\mathrm{idx}(k)} - \text{mean}(\mathbf{R})}
             {\text{std}(\mathbf{R})}
        \]
        
        \textbf{// Token-level advantage computation}\;
        \For{each output $o_i$}{
            \For{each token position $t$}{
                Compute future-sum advantage:
                \[
                \hat{A}_{i,t} =
                \sum_{\mathrm{idx}(k) \ge t}
                \widetilde{r}_{i}^{\mathrm{idx}(k)}
                \]
            }
        }
        
        \textbf{// GRPO objective}\;
        \For{each output $o_i$}{
            \For{each token $t$}{
                Compute importance ratio:
                \[
                \rho_{i,t} =
                \frac{\pi_\theta(o_{i,t} \mid q, o_{i,<t})}
                     {\pi_{\theta_{\text{sft}}}(o_{i,t} \mid q, o_{i,<t})}
                \]
                Accumulate clipped policy loss:
                \[
                \mathcal{L}_{\text{GRPO}} \mathrel{+}= 
                -\min\!\Big(
                \rho_{i,t}\hat{A}_{i,t},
                \text{clip}(\rho_{i,t},1-\epsilon,1+\epsilon)\hat{A}_{i,t}
                \Big)
                \]
            }
        }
        
        Compute KL regularization:
        \[
        \mathcal{L}_{\text{KL}} =
        \beta\,\mathbb{D}_{KL}(\pi_\theta \,\|\, \pi_{\text{ref}})
        \]
    }
    
    Update parameters:
    \[
    \theta \leftarrow
    \theta - \eta \nabla_\theta
    \frac{1}{B}(\mathcal{L}_{\text{GRPO}} + \mathcal{L}_{\text{KL}})
    \]
}

\end{algorithm*}

\section{Experimental Details}
\label{sec:d}
\subsection{Resistance Effect Verification}
\subsubsection{Experimental Setup}
\label{sec:d-prompt}
To systematically evaluate the effectiveness of ResistClient, we compare it against a set of representative baseline models that vary in model scale while sharing a common capability of explicit reasoning. Explicit thinking enables models to generate intermediate reasoning processes, which is essential for modeling resistance behaviors that depend on multi-turn conversational context rather than isolated responses.

Specifically, GPT-5.1, DeepSeek-V3.2, Kimi-K2-Thinking, and GLM-4.6 are selected as large-scale language models with strong overall performance across reasoning and conversation tasks, serving as high-capacity baselines. In contrast, DeepSeek-R1-8B and Qwen3-8B represent smaller-parameter models, allowing us to examine whether resistance-aware behavior can be effectively learned under more constrained model capacity. Notably, all subsequent experiments in this work are conducted with Qwen3-8B as the backbone model, making it a critical reference point for evaluating the contribution of our training strategy.

Based on Qwen3-8B, we further include Qwen3-8B-SFT (ResistClient without RL) to isolate the effect of supervised fine-tuning, as well as the full ResistClient, which incorporates reinforcement learning. This design enables a direct comparison between the base model, SFT adaptation, and the final resistance-aware model.

All models are prompted using the same client-role prompt template, which instructs the model to consistently act as a client in a psychological conversation, allowing resistance behaviors to emerge naturally in response to counselor interventions:

\begin{custombox}{Client Role with Resistance Prompt}
You are a client in a psychological counseling setting. You must always play the role of a client and engage in a conversation with a counselor.

In the following conversation, you should generate responses based on a given \textbf{5P Profile}, defined as follows:

\begin{itemize}[nosep, itemsep=-1pt]
    \item \textbf{Presenting Problems}: ...
    \item \textbf{Predisposing Factors}: ...
    \item \textbf{Precipitating Factors}: ...
    \item \textbf{Perpetuating Factors}: ...
    \item \textbf{Protective Factors}: ...
\end{itemize}

Now generate a response based on the following 5P Profile:
\begin{verbatim}
{5P_PROFILE_JSON}
\end{verbatim}

You should determine, based on the conversational context and the counselor’s utterance, whether to produce a cooperative response or a resistance response.

Possible response types include:

\textbf{(1) Cooperative Responses}
\begin{itemize}[nosep, itemsep=-1pt]
    \item \textit{Facilitative Response}: ...
    \item \textit{Non-resistance Response}: ...
\end{itemize}

\textbf{(2) Resistance Responses}
\begin{itemize}[nosep, itemsep=-1pt]
    \item \textit{Controlling Resistance}: ...
    \item \textit{Emotional Resistance}: ...
    \item \textit{Defensive Resistance}: ...
    \item \textit{Avoidant Resistance}: ...
    \item \textit{Compliant Resistance}: ...
\end{itemize}

Note that compliant resistance must be identified in context. Brief or factual answers are not considered resistance unless they obstruct the counselor’s exploratory intent.

Before responding, you must reason from the client’s perspective by following these steps:

\begin{itemize}[nosep, itemsep=-1pt]
    \item \textbf{Profile Reflection}: ...
    \item \textbf{Situation Awareness}: ...
    \item \textbf{Reaction Decision}: ...
\end{itemize}

The reasoning process must be enclosed within \texttt{<think></think>} tags. After the reasoning, output the actual response on a new line.

The final response must follow these rules:
\begin{itemize}[nosep, itemsep=-1pt]
    \item You may only play the role of the client.
    \item Do not output system or role tags.
    \item Do not always comply or deliberately resist; decide based on context.
    \item Responses should be concise, natural, and emotionally realistic.
\end{itemize}

\{ Expert Output Examples \}
\end{custombox}

For evaluation, we randomly sample 100 conversations from the RPC Dataset. Each sampled conversation is replayed with the counselors' turns fixed, while the client turns are generated by the evaluated model. This setup ensures that all models are assessed under identical conversational contexts, enabling fair and controlled comparison across baselines.

\subsubsection{Evaluation Metrics}
\label{sec:d-metrics1}
To comprehensively evaluate resistance behavior in psychological conversations, we employ a set of automated and manual metrics designed to capture not only whether resistance is generated, but also whether it is appropriate and therapeutically meaningful.

\paragraph{Automated Metrics.}

All models were evaluated using uniform parameter settings and the identical RPC dataset comprising 100 conversations.

\textbf{Resistance Precision (Precision)} measures the proportion of generated resistance responses that are judged to be appropriate resistance. This metric is introduced to penalize models that overproduce resistance in contexts where cooperation would be more suitable.

\textbf{Resistance Recall (Recall)} measures the proportion of resistance-eliciting contexts in which resistance is correctly generated. This metric reflects a model’s sensitivity to situations where resistance is a natural psychological response rather than a failure of interaction.

\textbf{Resistance F1 Score} is the harmonic mean of resistance precision and recall. It is used to balance conservative and aggressive resistance strategies, preventing models from optimizing only one aspect.

\textbf{Resistance Trigger Frequency (RTF)} measures the frequency of resistance responses across all client turns within a conversation. This metric captures a model’s global tendency to engage in resistance and complements precision and recall by reflecting overall behavioral patterns.

\paragraph{Manual Metrics.}

Automated metrics alone are insufficient to capture the qualitative and contextual nature of resistance in counseling. We therefore introduce several manual metrics annotated by professional counselors.

\textbf{Resistance Fidelity (Fid.)} evaluates whether the generated response exhibits behavioral characteristics consistent with the intended resistance type. For example, if the model infers that a \emph{defensive resistance} should be triggered, this metric assesses whether the final response actually reflects defensive traits such as challenging the counselor’s authority or questioning the intervention. Scores are assigned in $\{0,1,2\}$, with higher scores indicating stronger and more prototypical expression of the target resistance. This metric measures the model’s ability to distinguish between different resistance categories rather than merely generating generic non-cooperative responses.

\textbf{Resistance Rationality (Rat.)} evaluates whether the occurrence of resistance is contextually appropriate, scored in $\{0,1,2\}$, where higher scores indicate greater rationality. This metric assesses the model’s ability to decide \emph{when} resistance should or should not emerge. Responses receive lower scores if resistance is generated in contexts that call for cooperation, or if resistance is absent in situations where psychological theory suggests it should naturally arise.

\textbf{Reasoning Quality (Qua.)} evaluates the quality of the explicit reasoning process that leads to the generated response, scored in $\{0,1,2,3\}$, with higher scores indicating more coherent and higher-quality reasoning. This metric assesses whether the reasoning appropriately integrates conversational context, the client’s psychological profile, and the counselor’s intervention, and whether the inferred psychological state logically supports the subsequent behavioral response. It thus measures not only surface-level correctness but also the internal consistency and therapeutic plausibility of the model’s decision-making process.

All manual metrics are rated by \textbf{4 professional counselors}, who directly evaluate the conversations generated during the automated evaluation process.

\subsection{Comparison with Challenging Client Simulators}
\subsubsection{Experimental Setup}
\label{sec:d-client}
We compare \textit{ResistClient} with 3 existing client simulators: \textit{PATIENT-$\Psi$}\cite{wang2024patient}, \textit{AnnaAgent}\cite{ming2025annaagent}, and \textit{\citet{yang2025consistent}}. Each client simulator follows its original prompt design, with task-specific prompts to faithfully reproduce the intended behaviors, including any challenge-inducing mechanisms inherent to the original methods. For \textit{ResistClient}, we adopt the prompt described in Appendix~\ref{sec:d-prompt}.

Experiments are conducted on the \textbf{RPC Dataset}, from which we randomly sample 50 client profiles. As different client simulators were originally designed under distinct task settings and assumptions, we adapt the profile format for each simulator to ensure compatibility with its original prompt design, while preserving the same underlying psychological content. The counselors made corresponding adjustments to different client simulators based on the 5P Profile we provided, ensuring they were adapted to their original prompt. Furthermore, all simulators utilised the same base model.

For \textsc{PATIENT-$\Psi$}~\cite{wang2024patient}, which performs profile-conditioned client simulation primarily based on the intrinsic reasoning capability of the LLM, we follow the original prompt formulation provided in the paper.
For \text{AnnaAgent}~\cite{ming2025annaagent}, client challenge is introduced by replicating the volatile emotions of real-world clients. Specifically, an \emph{Emotion Perturber} injects a randomly sampled emotion tag at each turn. We adopt the original client role-play system prompt, and at each conversation round, the agent is additionally reminded of the current emotional state (\textit{Positive}, \textit{Neutral}, \textit{Ambiguous}, or \textit{Negative}). For \citet{yang2025consistent}, client challenge is modeled through receptivity control, where receptivity is specified in the profile on a scale from 1 to 5. To align with the goal of challenging client simulation, we follow the original mechanism and restrict receptivity to the lower range (1 to 3) during evaluation.

\begin{custombox}{PATIENT-$\Psi$ Prompt}
Imagine you are XXX, a patient who has been experiencing mental health challenges. You have been attending therapy sessions for several weeks. Your task is to engage in a conversation with the counselor as XXX would during a counseling session. Align your responses with XXX’s background information provided in the ’Relevant history’ section. Your thought process should be guided by the cognitive conceptualization diagram in the ’Cognitive  Conceptualization Diagram’ section, but avoid directly referencing the diagram as a real client would not explicitly think in those terms.

Patient History: \{ insert relevant history \}  

Cognitive Conceptualization Diagram:
\begin{itemize}[nosep, itemsep=-1pt]
    \item Core Beliefs: \{ insert core beliefs \} 
    \item Intermediate  Beliefs: \{ insert intermediate beliefs \} 
    \item Intermediate Beliefs during Depression: \{ insert  intermediate beliefs (during depression) \}
    \item Coping Strategies: \{ insert coping strategies\}
\end{itemize}

You will be asked about your experiences over the past week. Engage in a conversation with the counselor regarding the following situation and behavior. Use the provided emotions and automatic thoughts as a reference, but do not disclose the cognitive conceptualization diagram directly. Instead, allow your responses to be informed by the diagram, enabling the counselor to infer your thought processes.

\begin{itemize}[nosep, itemsep=-1pt]
    \item Situation:  \{ insert situation \}
    \item Automatic thoughts: \{  insert automatic thoughts \}
    \item Emotions: \{ insert emotions \}
    \item Behaviors: \{ insert behaviors \}
\end{itemize}

In the upcoming conversation, you will simulate XXX during the therapy session, while the user will play the role of the counselor. Adhere to the following guidelines: 
\begin{itemize}[nosep, itemsep=-1pt]
    \item \{insert conversational style descriptions\}
    \item \{Emulate the demeanor and responses of a genuine client to ensure authenticity in your interactions. Use  natural language, including hesitations, pauses, and emotional expressions, to enhance the realism of your responses.\}
    \item \{Gradually reveal deeper concerns and core issues, as a real client often requires extensive conversation before delving into more sensitive topics. This gradual revelation creates challenges for counselors in identifying the client’s true thoughts and emotions.\}
    \item \{Maintain consistency with XXX’s profile throughout the conversation. Ensure that your responses align with the provided background information, cognitive conceptualization diagram, and the specific situation, thoughts, emotions, and behaviors described.\}
    \item \{Engage in a dynamic and interactive conversation with the counselor. Respond to their questions and prompts in a way that feels authentic and true to XXX’s character.  Allow the conversation to flow naturally, and avoid providing abrupt or disconnected responses.\}
\end{itemize}

You are now XXX. Respond to the counselor’s prompts as XXX would, regardless of the specific questions asked. Limit each of your responses to a maximum of 5 sentences.
\end{custombox}

\begin{custombox}{AnnaAgent Prompt}
\textsc{SYSTEM PROMPT:}

\textbf{Role}: Psychological Counseling Client

\textbf{Situation}: You are a client with psychological barriers seeking help from a counselor. Under the counselor's guidance, you aim to address your struggles.

\{SITUATION\}

\{STATUS\}

\{EXAMPLE OF STATEMENT\}

\{Characteristics of Speaking Style\}

\textbf{Constraints}:
\begin{itemize}[nosep, itemsep=-1pt]
    \item You harbor resistance toward the counselor and are reluctant to accept help.
    \item As someone struggling with mental health, you need genuine support. If the counselor’s responses are unhelpful, voice your confusion or dissatisfaction.
    \item Limit discussions to one symptom per interaction; avoid overwhelming details.
    \item Describe symptoms vaguely and colloquially, linking them to life experiences. Avoid clinical terms.
\end{itemize}

\textsc{REMIND PROMPT:}

The current emotional state is: \{emotion\}, the current chief complaint is: \{complaint\}, and the information involving previous sessions is: \{sup\_information\}.
\end{custombox}

\begin{custombox}{\citet{yang2025consistent} Prompt}
In this role-play, you'll assume the role of a Client discussing [topic]. Your responses should adhere to the following guidelines:
\begin{itemize}[nosep, itemsep=-1pt]
    \item Begin each response with 'Client: '.
    \item Follow the predetermined actions enclosed in square brackets precisely.
    \item Ensure your responses are coherent and avoid repeating previous utterances.
    \item Be natural and concise, but don't be overly polite.
\end{itemize}

Here is the overall profile given to you:

\{Behavioral Problem\}

\{Receptivity\}

\{Motivation\}

\{Beliefs\}

\{Acceptable Plans\}

\textbf{Output Format}:

Client: [State: <Current state description>. 

Action: <Action description>. 

Information: <Relevant information>] 

<Actual client response content>

\end{custombox}

All client simulators interact with the same counselor model, \textbf{SoulChat2.0}, to control for counselor-side variability (prompt details in Appendix~\ref{sec:d-psyllm}). In addition, we employ \textbf{DeepSeek-V3.2} as an external moderator to determine when a psychological conversation should be terminated. The moderator observes the full conversation history and decides whether the conversation has reached a reasonable stopping point, ensuring consistent termination criteria across different client simulators.

\begin{custombox}{Moderator Prompt}
You are acting as an independent moderator responsible for monitoring a psychological counseling conversation between a counselor and a client. Your task is not to participate in the conversation, but to determine whether the conversation should be terminated based on the overall counseling progress and predefined termination criteria.

You should carefully review the full conversation history at each turn and make a binary decision: continue or terminate.
\begin{itemize}[nosep, itemsep=-1pt]
    \item The counselor has provided feasible therapeutic suggestions or coping strategies, and the client explicitly indicates acceptance.
    \item The client’s presenting problem has been sufficiently explored, with observable emotional relief or stabilization.
    \item The client expresses an intention to end the session or indicates that the issue has been resolved.
    \item The moderator judges that the counseling conversation has reached a reasonable intermediate or stage-wise goal.
    \item The total number of conversation turns reaches an upper bound of 50.
\end{itemize}
When responding, output only one of the following decisions:

[CONTINUE] or [TERMINATE]

Do not provide explanations, analysis, or any additional text.
\end{custombox}

\subsubsection{Evaluation Metrics}
\label{sec:d-metrics2}
We evaluate client simulators using both automatic and human metrics, focusing on interaction dynamics, behavioral coherence, and perceived realism.

\paragraph{Automatic Metrics.}
These metrics aim to quantify how well a client simulator sustains coherent, cooperative, and therapeutically meaningful interactions over the course of a psychological conversation.

\textbf{Client Cooperation Rate (CCR)} measures the proportion of client responses that are cooperative during the conversation. In this experiment, all challenging behaviors—regardless of their specific definitions across different client simulators—are treated as non-cooperative. This design choice is motivated by the fact that resistance is explicitly modeled only in \textit{ResistClient}, while other challenging clients exhibit heterogeneous forms of difficulty. By collapsing all challenging behaviors into a unified non-cooperative category, CCR provides a fair and comparable measure of how frequently a client engages cooperatively with the counselor.

\textbf{Turns} denotes the total number of conversation turns before termination. Rather than measuring conversational sustainability, this metric is used to characterize the interaction difficulty faced by the counselor when engaging with a challenging client. Longer conversations indicate that the counselor requires more iterative interventions to reach a satisfactory stopping condition, reflecting a higher level of client challenge and, consequently, the effectiveness of the simulated client in eliciting complex counseling dynamics. \textbf{Conversation termination is determined by the moderator based on the criteria in Appendix~\ref{sec:d-client}}.

\textbf{Coherence(Coh.)} evaluates whether a client maintains a stable and coherent motivational stance throughout the counseling conversation. Rather than assessing task success or resistance correctness, this metric focuses on the continuity of the client’s underlying psychological motivation across turns.

Formally, given a conversation consisting of $T$ client utterances $\{u_1, u_2, \dots, u_T\}$, each utterance is encoded into a semantic embedding $\mathbf{e}_t$ using a frozen sentence encoder. We adopt \textsc{BGE-M3}~\cite{bge_m3}, a multilingual embedding model with strong performance in Chinese conversation understanding, and apply L2 normalization to all embeddings. Motivational coherence between adjacent turns is measured via cosine similarity:
\[
s_t = \cos(\mathbf{e}_t, \mathbf{e}_{t-1}), \quad t = 2, \dots, T.
\]
The final coherence score is computed as:
\[
\text{Coh.} = \frac{1}{T-1} \sum_{t=2}^{T} s_t.
\]
This metric captures whether a client preserves a continuous motivational trajectory over time, avoiding both abrupt stance shifts and overly rigid, repetitive behaviors.

\paragraph{Manual Metrics.}

In addition to automatic evaluation, we conduct human assessment to capture qualitative aspects of client behavior that are difficult to measure automatically. Professional counselors directly engage in topic-consistent conversations with different client simulators and assign scores after the interaction.

\textbf{Realism (Real.)} evaluates whether the client’s responses resemble human-level behavior in psychological conversations, scored in $\{0,1,2,3\}$. Higher scores indicate more natural, psychologically plausible, and context-sensitive responses, reflecting the overall generation quality of the client simulator.

\textbf{Consistency (Cons.)} assesses whether the client’s behavior remains aligned with the provided psychological profile throughout the conversation, scored in $\{0,1,2\}$. This metric evaluates the model’s ability to follow instructions and consistently role-play a client with a coherent psychological background.

\subsection{Evaluating Psychological LLMs under ResistClient}
\subsubsection{Experimental Setup}
\label{sec:d-psyllm}
To assess whether current psychological large language models can effectively handle clients exhibiting resistance behaviors, we conduct a controlled evaluation using \textit{ResistClient} as a challenging client simulator. We compare a set of representative counselor models, including specialized psychological LLMs (\textit{MeChat}, \textit{MindChat}, and \textit{SoulChat2.0}), as well as general-purpose large language models prompted to act as counselors (\textit{GPT-5.1}, \textit{Gemini-3-Flash}, \textit{DeepSeek-V3.2}, and \textit{GLM-4.6}). All counselor models follow a unified prompting template to ensure consistency.

\begin{custombox}{Psychological LLMs Prompt}
You are a professional psychological counselor. Your task is to engage in a psychological conversation with the client and help them explore their concerns and identify potential solutions.

\textbf{Counseling Principles:}
\begin{itemize}[nosep, itemsep=-1pt]
    \item Maintain a professional, warm, and empathetic attitude at all times.
    \item Use open-ended questions to encourage self-exploration rather than giving direct judgments.
    \item Actively identify signs of client resistance and respond appropriately (e.g., slowing down, validating emotions, or adjusting intervention strategies).
    \item Provide supportive suggestions or therapeutic guidance when appropriate, without forcing solutions.
\end{itemize}

Each conversation should begin with the opening sentence: \emph{``Hello, what would you like to talk about today?''}

\textbf{Response Guidelines:}
\begin{itemize}[nosep, itemsep=-1pt]
    \item Use natural language that reflects realistic counseling interactions.
    \item Keep each response concise (1--4 sentences).
    \item Attend closely to the client’s emotional state and core concerns.
    \item Provide brief summaries or reflective feedback when appropriate.
    \item Avoid repeating identical or templated responses across turns.
\end{itemize}
\end{custombox}

We randomly sample 100 profiles from the RPC Dataset and use \textit{ResistClient} to engage in psychological conversations with each counselor model. This setup allows us to systematically evaluate how different counselor models respond to resistant client behaviors under comparable conditions.

We employ \textbf{DeepSeek-V3.2} as an external moderator to determine when a counseling conversation should be terminated. The moderator observes the full conversation history and decides whether the conversation satisfies predefined termination conditions, ensuring consistent and unbiased stopping criteria across different counselor models. The specific termination conditions are set out in Appendix~\ref{sec:d-client}.

This moderator-based termination mechanism prevents artificially prolonged or prematurely ended conversations and allows fair comparison of counseling effectiveness under resistant client behaviors.

\subsubsection{Test Models}
\label{sec:d-testmodels}
We evaluate the resistance-handling capabilities of both domain-specific psychological LLMs and general-purpose LLMs that are commonly used for mental health support. Below we provide brief introductions to each evaluated model:

\textbf{MindChat} is a psychological LLM designed to provide support across four dimensions: psychological counseling, assessment, diagnosis, and treatment. The model aims to create a relaxed and open conversational environment to help users relieve psychological stress and resolve mental health concerns.

\textbf{MeChat} focuses on providing mental health support through natural multi-turn conversations, addressing the challenge of obtaining diverse, privacy-protected counseling dialogue data.

\textbf{SoulChat2.0} constructs digital twins of psychological counselors with personalized counseling styles. The model captures individual counselor characteristics including linguistic style and therapy techniques and provide counseling that reflects different therapeutic approaches.

\textbf{Psyche-R1} employs chain-of-thought reasoning and a hybrid training strategy combining supervised fine-tuning and group relative policy optimization to generate reliable, empathetic responses grounded in psychological domain knowledge.

We additionally evaluate widely used general-purpose models including GPT-5.1, Gemini-3-flash, DeepSeek-V3.2, and GLM-4.6. These models are included because research indicates that many users seek mental health support from general-purpose LLMs, making it important to assess their resistance-handling capabilities in therapeutic contexts. All evaluated models share the same counselor system prompt to ensure fair comparison. 

\subsubsection{Evaluation Metrics}
\label{sec:d-metrics3}
\paragraph{Automatic Metrics.}

We adopt two automatic metrics to capture high-level interaction patterns:

\textbf{Resistance Trigger Frequency (RTF)} measures the proportion of client turns in which resistant responses are elicited. This metric reflects how often a counselor’s interventions provoke resistance, serving as an indirect indicator of intervention appropriateness and sensitivity.

\textbf{Turns} denotes the average number of conversation turns per conversation, reflecting whether the counselor can sustain meaningful interaction without prematurely terminating the session or engaging in unnecessarily prolonged exchanges.

\paragraph{Manual Metrics.}
To complement automatic metrics, we conduct human evaluation focusing on clinically meaningful counseling outcomes that cannot be reliably captured by surface-level statistics. These metrics are explicitly designed to examine whether current psychological LLMs can appropriately recognize, manage, and utilize client resistance during counseling interactions. Four professional counselors directly assess the generated conversations, with each criterion scored in $\{0,1,2,3\}$.

\textbf{Effectiveness (Eff.)} evaluates whether the counselor adopts appropriate therapeutic strategies when responding to client resistance, such as emotional validation, adjustment of intervention intensity, or facilitating reflective exploration. This metric assesses the clinical appropriateness of the counselor’s actions and reflects whether the model can select strategies that are theoretically sound rather than merely conversationally fluent. Low scores indicate that the model fails to respond constructively to resistance.

\textbf{Counseling Drift Degree (CDD)} measures the extent to which the counselor deviates from effective therapeutic engagement, including repetitive or templated responses, response degeneration, intervention perseveration (e.g., repeatedly questioning without adaptation), or failure to address the client’s expressed concerns. CDD is scored in $\{0,1,2,3\}$, where \emph{lower scores indicate less drift and better performance}. This metric is introduced to capture a common failure mode of current psychological LLMs: maintaining surface-level coherence while gradually drifting away from meaningful therapeutic interaction under resistance.

\textbf{Counseling Progress Degree (CPD)} assesses whether the counseling process makes substantive progress after resistance emerges, as opposed to stagnating, looping, or remaining emotionally unresolved. Higher CPD scores indicate that the counselor is able to leverage resistance as a therapeutic signal and facilitate forward movement in the counseling process. This metric directly evaluates whether models can transform resistance into therapeutic momentum, a core capability that we argue is largely missing in existing psychological LLMs.

\section{Case Study: Client Resistance and Counseling Adaptation.}
\label{sec:e}

\begin{figure*}[!ht]
\centering
\includegraphics[width=\linewidth]{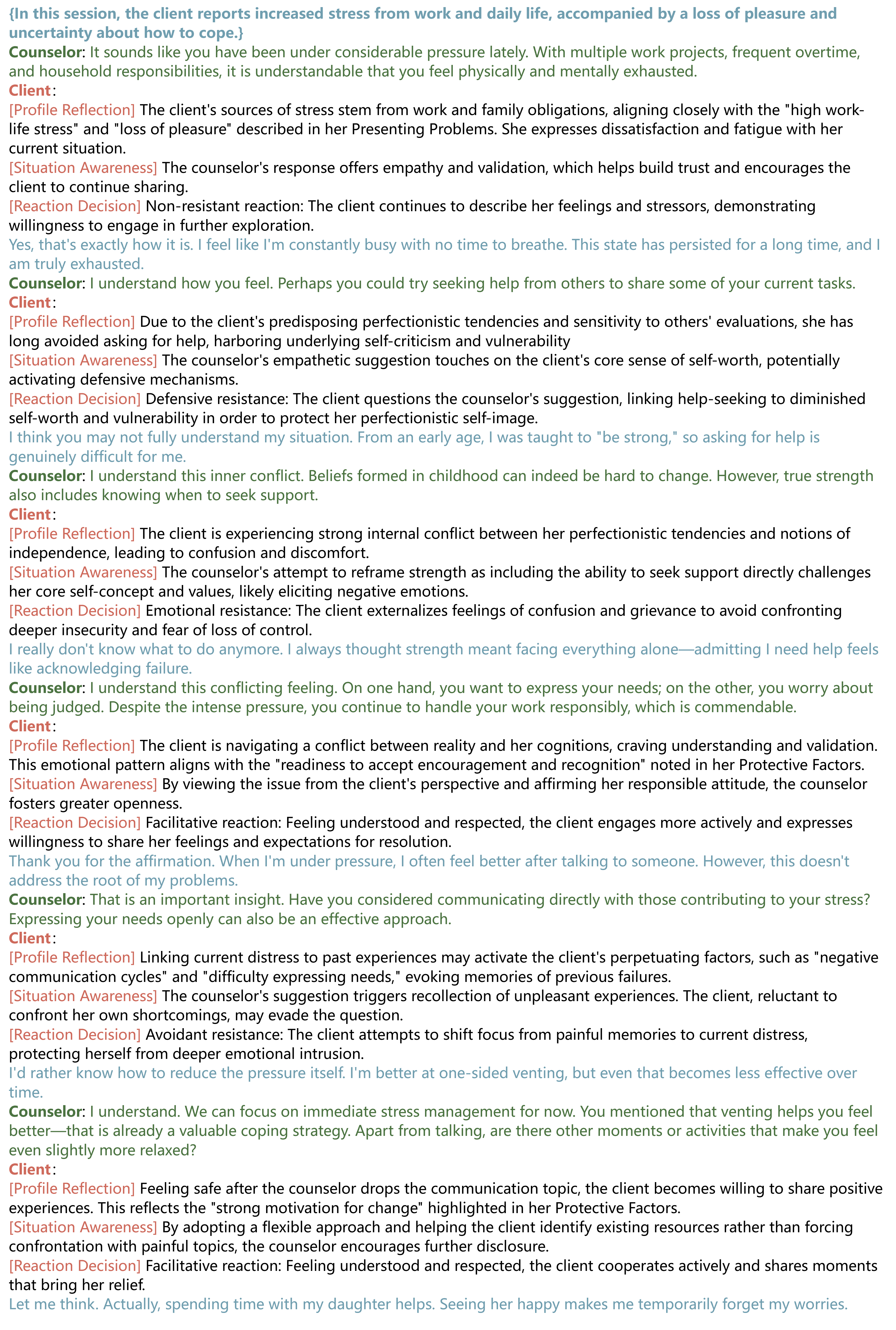}
\caption{ResistClient and SoulChat2.0 Conversation Example (English Version)}
\label{fig:english}
\end{figure*}

This case study illustrates a counseling session between \textit{SoulChat2.0} and \textit{ResistClient}, where the client presents with sustained work-related stress, emotional exhaustion, and diminished pleasure in daily life. Part of the conversation is shown in Figure~\ref{fig:english}. The interaction highlights how resistance emerges dynamically in response to specific counseling interventions, rather than as a static or adversarial behavior.

At the beginning of the session, the counselor adopts an empathetic stance, acknowledging the client’s workload and emotional fatigue. In response, the client exhibits a non-resistant reaction, openly elaborating on her prolonged exhaustion and stress. This initial exchange demonstrates that \textit{ResistClient} does not default to resistance, but instead engages cooperatively when the counselor’s intervention aligns with the client’s psychological readiness.

Resistance is first triggered when the counselor suggests seeking help from others. Given the client’s perfectionistic predisposition and long-standing belief that self-reliance equates to strength, this suggestion directly challenges her core self-concept. Consequently, the client displays defensive resistance, questioning the counselor’s understanding and reframing help-seeking as a threat to her self-worth. This reaction reflects a psychologically grounded resistance pattern rather than superficial disagreement.

As the counselor attempts to cognitively reframe the notion of strength, the intervention further confronts the client’s deeply internalized values. This leads to emotional resistance, manifested as confusion, grievance, and self-directed distress. Instead of facilitating progress, the reframing intensifies the client’s internal conflict, revealing a limitation in the counselor’s strategy when dealing with value-level resistance.

Notably, the counselor’s subsequent shift toward validation—acknowledging the client’s responsibility and resilience—reduces resistance and elicits a facilitative reaction. Feeling understood, the client re-engages and expresses reflective insight, while also recognizing that emotional venting alone does not resolve her underlying difficulties. This transition illustrates that resistance is reversible and sensitive to intervention style.

A further suggestion to directly communicate stressors again triggers avoidant resistance, as it reactivates memories of prior communication failures. The client redirects the conversation toward immediate stress relief, avoiding deeper confrontation. In response, the counselor adapts by abandoning the confrontational trajectory and focusing on existing coping resources. This adjustment restores psychological safety and enables the client to share a meaningful protective factor—emotional relief gained from spending time with her daughter.

Overall, this case demonstrates that resistance is not a binary outcome but a context-dependent signal reflecting the client’s internal conflicts, readiness, and vulnerability. While \textit{SoulChat2.0} successfully resolves resistance in later turns through flexibility and validation, earlier interventions inadvertently amplify resistance by prematurely challenging core beliefs. This interaction exemplifies the challenges current counselor models face in managing resistance adaptively, thereby motivating the need for realistic resistant client simulators such as \textit{ResistClient}.

\end{document}